\documentclass{article}
\usepackage{iclr2027_conference,times}

\usepackage{amsmath,amssymb}
\usepackage{algorithm}
\usepackage{algpseudocode}
\usepackage{graphicx}
\usepackage{subcaption}
\usepackage{booktabs}
\usepackage{array}
\usepackage{url}
\usepackage{xspace}
\usepackage{xcolor}
\usepackage[hidelinks]{hyperref}

\newsavebox{\iclrtablebox}
\newenvironment{fittoiclr}{\begin{lrbox}{\iclrtablebox}}{%
  \end{lrbox}%
  \ifdim\wd\iclrtablebox>\linewidth
    \resizebox{\linewidth}{!}{\usebox{\iclrtablebox}}%
  \else
    \usebox{\iclrtablebox}%
  \fi
}

\newif\ifrevisionhighlight
\revisionhighlightfalse
\newcommand{\rev}[1]{\ifrevisionhighlight\textcolor{red}{#1}\else#1\fi}
\newenvironment{revblock}{\ifrevisionhighlight\color{red}\fi}{}
\newcommand{\revb}[1]{\ifrevisionhighlight\textcolor{blue}{#1}\else#1\fi}

\definecolor{pruningrevisionblue}{HTML}{174EA6}
\newif\ifpruningrevisionhighlight
\pruningrevisionhighlightfalse
\newcommand{\pruningrev}[1]{\ifpruningrevisionhighlight\textcolor{pruningrevisionblue}{#1}\else#1\fi}
\newcommand{\relatedrev}[1]{\pruningrev{#1}}
\newcommand{\experimentrev}[1]{\pruningrev{#1}}
\newif\iffullversion
\fullversiontrue
\newcommand{\fv}[2]{\iffullversion#1\else#2\fi}
\providecommand{\Description}[1]{}
\providecommand{\FloatBarrier}{\clearpage}

\title{Agentic Search for Counterfactual Recourse\\ under Fixed LLM Budgets}
\author{Yasuo Tabei\\
RIKEN Center for Advanced Intelligence Project\\
Tokyo, Japan\\
\texttt{yasuo.tabei@riken.jp}}
\iclrfinalcopy
\hypersetup{
  pdftitle={Agentic Search for Counterfactual Recourse under Fixed LLM Budgets},
  pdfauthor={Yasuo Tabei}
}

\begin{document}
\maketitle
\lhead{Preprint}

\begin{abstract}
Counterfactual recourse aims to provide actionable feature changes that would alter an unfavorable decision made by a predictive model. In practice, affected individuals often benefit from multiple feasible alternatives rather than a single optimal explanation. \rev{A natural way to produce such alternatives is to prompt large language models (LLMs). However, prompting incurs a practical constraint:} the number of LLM calls is often the dominant computational and economic cost. %
\rev{Together, the need for multiple alternatives and this cost constraint shift the problem from finding a single high-quality counterfactual to efficiently generating a set of oracle-validated counterfactuals under a fixed LLM-call budget.}
In this work, we study counterfactual recourse generation in the LLM-agentic setting as a fixed-budget search problem and propose Recourse Monte Carlo Tree Search (ReCo-MCTS), an agentic tree-search framework \pruningrev{that aims to increase the yield of unique, oracle-validated counterfactuals under this budget while accounting for the extent of the required changes}.
\pruningrev{ReCo-MCTS combines LLM-based multi-candidate generation, constraint checking, black-box oracle evaluation, and UCT-guided tree search to accumulate valid counterfactuals under a fixed LLM-call budget.}
\experimentrev{On four real-world tabular datasets, ReCo-MCTS returns more counterfactuals than the evaluated baselines in our main comparison under the specified resource limits, with trade-offs in the extent of the required changes.}

\end{abstract}

\section{Introduction}

Autonomous decision systems are increasingly popular in high-stakes domains such as credit approval, hiring, and healthcare.
\rev{For example, automated credit scoring can reject loan applications, and algorithmic hiring systems can filter out candidates.}
Across these domains, the key question for affected individuals is not only why the decision was made but also what feasible changes could lead to a desired outcome.
One natural way to answer \revb{this} is through counterfactual explanations, a concrete form of actionable recourse that recommends feature changes \rev{leading} to a desired model prediction~\citep{wachter2018counterfactual,ustun2019actionable}.%

\begin{figure}[t]
  \centering
  \includegraphics[width=0.8\linewidth, trim=7 36 7 23, clip]{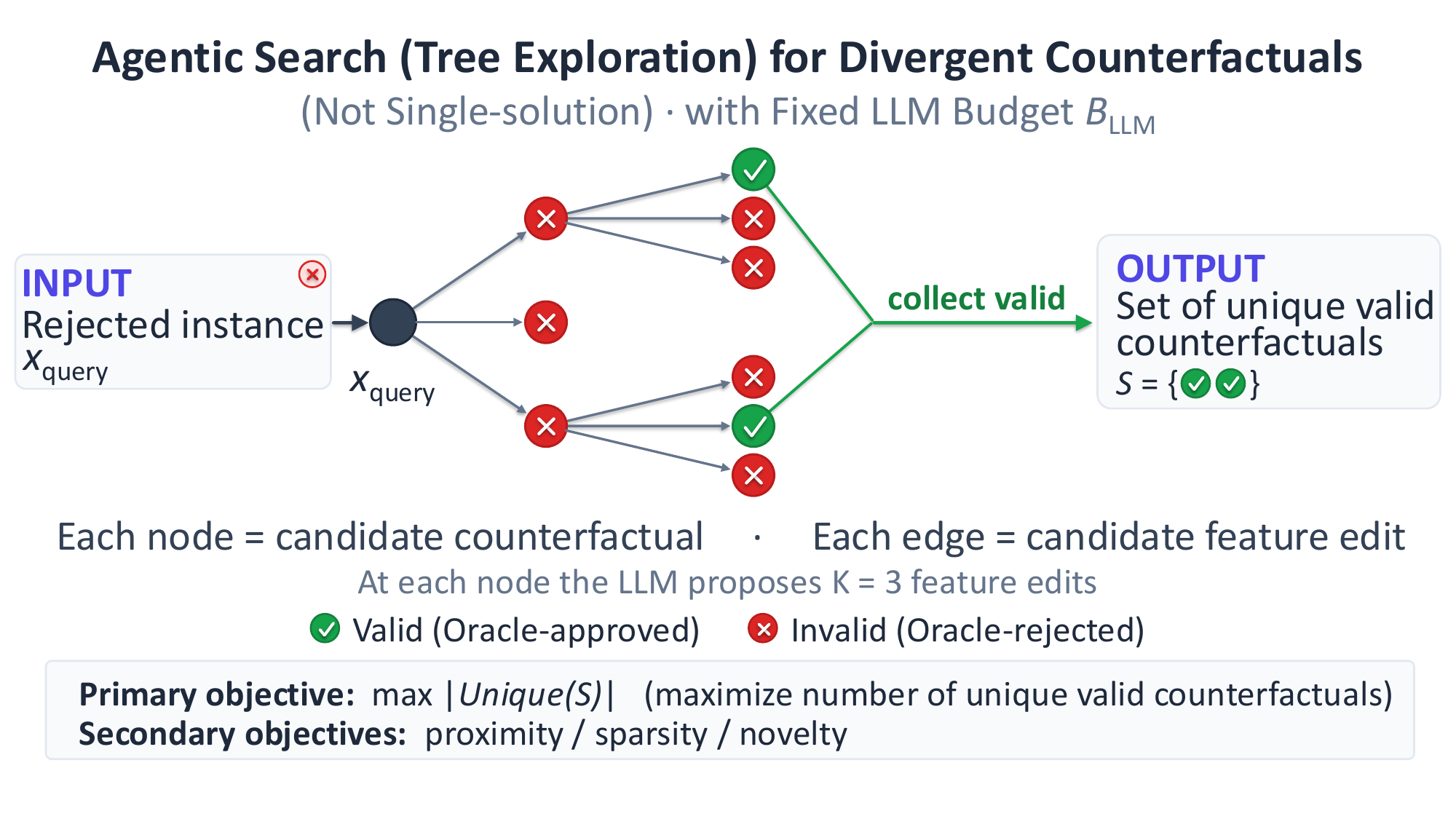}%
  \vspace{-0.0cm}
  \caption{\pruningrev{\textbf{Illustration of counterfactual recourse as fixed-budget search.}
}}\label{fig:comp_mcts_overview}
  \Description{A tree of candidate feature states rooted at a rejected input, with oracle-approved and oracle-rejected nodes and a returned set of valid counterfactuals. Three candidate proposals per expansion are shown for illustration.}
\end{figure}

Early counterfactual explanation methods cast recourse as a distance-minimization problem~\citep{wachter2018counterfactual,ustun2019actionable}.
\rev{However, in certain regimes proximity-driven objectives become structurally close to adversarial example generation, yielding boundary-crossing edits that are often not semantically meaningful~\citep{pawelczyk22a}.}
\rev{Moreover, many existing counterfactual methods still return only one or a small number of proximity-driven solutions, whereas users often benefit from multiple feasible alternatives.}
\rev{This gap between single-solution methods and users' need for alternatives} suggests that counterfactual recourse generation should be viewed not as a single-solution optimization problem, but as a search problem that aims to produce a diverse set of valid alternatives.

\looseness=-1 %
\pruningrev{We study a \emph{training-free, black-box} setting: search queries the decision model---the \emph{oracle}---without accessing its internals or training rows, but uses supplied feature constraints and training-derived distance scales. No recourse generator is trained~\citep{verma2022amortized,vo2023feature}.
This setting applies when training rows cannot be shared but feature metadata and summary statistics are available.
Off-the-shelf LLMs can propose varied edits over mixed numerical and categorical features without task-specific training.}

\rev{Such LLM proposals, however, are not guaranteed to achieve the desired model prediction, so each candidate must be strictly validated by the oracle; moreover, under cost and latency constraints, the number of LLM calls is the primary practical budget.}
\rev{An important open challenge is therefore to design a fixed-budget search method that maximizes the yield of \emph{unique, oracle-validated} counterfactuals---rather than converging to near-duplicate solutions---while maintaining favorable quantity--quality trade-offs.}

{\em Contributions.} We propose \textbf{Recourse Monte Carlo Tree Search (ReCo-MCTS)}, \rev{an agentic tree search method} based
on Monte Carlo Tree Search (MCTS)~\citep{browne2012survey,kocsis2006bandit}~(Figure~\ref{fig:comp_mcts_overview}).
\pruningrev{The key idea is to generate multiple candidate feature edits per LLM call and accumulate valid counterfactuals across explored branches. ReCo-MCTS combines constraint checking, black-box oracle evaluation, and UCT-guided tree search to support this process under a fixed LLM-call budget.}
Our contributions are as follows:
\begin{list}{\arabic{enumi}.}{%
  \usecounter{enumi}%
  \setlength{\leftmargin}{1.6em}%
  \setlength{\labelwidth}{1.6em}%
  \setlength{\labelsep}{0.3em}%
  \setlength{\itemsep}{0.2em}%
  \setlength{\topsep}{0.2em}%
}
\item {\bf Problem formulation:} \rev{We formulate counterfactual recourse generation as a fixed-budget search problem that aims to maximize, under a fixed LLM-call budget, the yield of \emph{unique, oracle-validated} counterfactuals while maintaining favorable quantity--quality trade-offs, rather than converging to a single best counterfactual.}
\item {\bf Search framework:} \pruningrev{We propose \textbf{Recourse Monte Carlo Tree Search (ReCo-MCTS)}, a budget-aware agentic tree search method based on Monte Carlo Tree Search (MCTS)~\citep{browne2012survey,kocsis2006bandit} that combines LLM-based multi-candidate generation, constraint checking, and black-box oracle evaluation, using candidate rewards to guide subsequent tree selection.}
\item {\bf Empirical findings:} \experimentrev{On four tabular datasets, increasing the number of candidates per call raises CF count under 30 LLM calls. Allowing up to three edits yields more CFs than single-edit proposals, while single-edit proposals require fewer changed features. In the main comparison with DiCE, Growing Spheres, and C-CHVAE, ReCo-MCTS has higher CF counts, whereas DiCE changes fewer features and has lower recorded search time. The methods use different resource budgets, as detailed in Section~\ref{sec:experiments}.}
\end{list}

\section{Related Work}
\begin{table}[h]
    \caption{Summary of representative counterfactual explanation methods.}
    \vspace{-0.2cm}
    \label{tab:comparison_methods}
    \centering
    \fontsize{8}{9}\selectfont
    \setlength{\tabcolsep}{3pt}
    \renewcommand{\arraystretch}{1.10}
    {\ifpruningrevisionhighlight\color{pruningrevisionblue}\fi
    \begin{tabular}{@{}
      >{\raggedright\arraybackslash}p{\dimexpr0.24\linewidth-1.5\tabcolsep\relax}
      >{\raggedright\arraybackslash}p{\dimexpr0.245\linewidth-1.5\tabcolsep\relax}
      >{\raggedright\arraybackslash}p{\dimexpr0.265\linewidth-1.5\tabcolsep\relax}
      >{\raggedright\arraybackslash}p{\dimexpr0.25\linewidth-1.5\tabcolsep\relax}@{}}
    \toprule
    \textbf{Method} & \textbf{Primary Objective} & \textbf{Generation / Search} & \textbf{Returned CFs} \\
    \midrule
    \citet{wachter2018counterfactual} & Distance minimization & Loss-based optimization & One CF per optimization \\
    \addlinespace[2pt]
    Growing Spheres~\citep{laugel2018comparison} & Proximity and sparsity & Sphere sampling; sparse refinement & One CF per search \\
    \addlinespace[2pt]
    DiCE~\citep{mothilal2020dice} & Proximity and set diversity & Joint CF-set optimization & A requested CF set \\
    \addlinespace[2pt]
    CERTIFAI~\citep{sharma2020certifai} & Distance minimization & Genetic search & Requested number selected by fitness \\
    \addlinespace[2pt]
    C-CHVAE~\citep{pawelczyk2020learning} & Proximity in data-supported regions & Learned latent-space sampling & One CF per search \\
    \midrule
    \textbf{ReCo-MCTS} (this work) & \textbf{Distinct valid CF yield under a fixed LLM-call budget} & \textbf{Multiple LLM proposals per call; UCT-guided search} & \textbf{All valid CFs accumulated across explored branches} \\
    \bottomrule
    \end{tabular}
    }
\vspace{-0.5cm}    
\end{table}
\looseness=-1 %
\relatedrev{Table~\ref{tab:comparison_methods} contrasts proximity- and set-oriented objectives with ReCo-MCTS's focus on distinct valid CF yield under a fixed LLM-call budget. Entries describe cited formulations; experimental variants and repeated-return collection appear in Section~\ref{sec:comparison_existing}. Detailed reviews are in \fv{Appendix~\ref{app:related_work_details}}{the full version}.}

\subsection{Distance-Minimization and the Adversarial Trap}
Early approaches formulated recourse as a distance-minimization problem, seeking the closest counterfactual $x'$ to the query $x_{\text{query}}$ under an input-space distance~\citep{wachter2018counterfactual,ustun2019actionable}.
\revb{Methods such as Growing Spheres~\citep{laugel2018comparison} and DiCE~\citep{mothilal2020dice} extend this formulation with sparsity and diversity objectives, and MOC~\citep{dandl2020multi} uses multi-objective optimization to expose trade-offs. CERTIFAI~\citep{sharma2020certifai} applies genetic-algorithm search for robustness and fairness analyses, yet remains proximity-driven.} 
However, proximity-driven objectives can be vulnerable to the ``adversarial trap''~\citep{pawelczyk22a}, where boundary-crossing edits exploit model-specific vulnerabilities rather than \rev{corresponding to} semantically meaningful changes.
Moreover, these methods typically do not enforce oracle-validated validity as a hard constraint and can spend much of a limited budget on near-duplicate candidates.

\subsection{Generative and Structured Search for Recourse}
\rev{Recent works utilize generative models to improve the plausibility of recourse.}
Manifold-based methods like C-CHVAE~\citep{pawelczyk2020learning} and diffusion-based frameworks (e.g., SCD)~\citep{satml24} generate realistic samples by modeling the data distribution.%
\rev{However, C-CHVAE focuses on the single closest counterfactual. Sampling-based approaches such as SCD do not formulate generation as a budgeted search process and typically lack explicit redundancy control, which limits the yield of distinct valid options.}

\rev{A different line of work computes recourse as minimal-cost \emph{interventions} under a specified Structural Causal Model (SCM)~\citep{karimi2021algorithmic}.
While compelling when a reliable causal model is available, these approaches typically require at least a known causal graph and additional structural assumptions, which are often unavailable in strictly black-box settings.}

\begin{revblock}
Complementary to the model-based generators above, another line of work \rev{obtains diverse recourse through learned generators or explicit optimization}: amortized methods train a model that produces sequential recourses for black-box models with lightweight inference at test time~\citep{verma2022amortized}; feature-based learning yields diverse and privacy-preserving counterfactuals~\citep{vo2023feature}; and mixed-integer formulations efficiently search for diverse coherent explanations~\citep{russell2019efficient}.
These approaches are attractive when training data and an offline training phase are available, but they require access to the data distribution and typically need re-training when the underlying model changes.
\relatedrev{ReCo-MCTS requires no recourse-generator training or access to classifier internals or training rows during search. It uses supplied feature constraints and training-derived distance scales, complementing learned generators.}
\end{revblock}

LLM-based approaches have also emerged. \revb{While zero-shot LLMs can generate plausible counterfactuals~\citep{bhattacharjee2024zeroshot}, they act mainly as flexible proposal generators, lacking systematic exploration, strict oracle validation, and explicit budget control.} %

\relatedrev{Despite substantial progress in counterfactual generation and agentic search, to the best of our knowledge no prior work in the LLM-agentic recourse setting has explicitly targeted a high yield of unique, oracle-validated counterfactuals under a fixed LLM-call budget while examining the associated quantity--quality trade-offs. To address this gap, we formulate counterfactual recourse generation as a fixed-budget search problem.}
\pruningrev{We then present ReCo-MCTS, an MCTS framework that combines LLM-based multi-candidate generation, constraint checking, black-box oracle evaluation, and UCT-guided tree search to accumulate valid counterfactuals across explored branches.} Details of our proposed method are presented in the next section.

\section{Method}
\label{sec:method}

\subsection{Problem Formulation}
\label{sec:problem_formulation}
Let $\mathcal{X} \subseteq \mathbb{R}^D$ denote the input space, and let $f: \mathcal{X} \to \{0,1\}$ be a black-box classifier (oracle), where $f(x)=1$ denotes \emph{approved} and $f(x)=0$ denotes \emph{rejected}. Given a rejected instance $x_{\text{query}} \in \mathcal{X}$ with $f(x_{\text{query}})=0$, our goal is to generate a set of counterfactuals $\mathcal{S}$ under a fixed LLM-call budget $B_{\mathrm{LLM}}$ such that every \revb{$x' \in \mathcal{S}$} is approved by the oracle, i.e., $f(x')=1$.%
We restrict candidates to feasible interventions on the query instance: let $\mathcal{A}(x_{\text{query}})\subseteq\mathcal{X}$ denote the set of allowable counterfactuals that respect actionability/immutability and feature-domain constraints (dataset-specific).

Our objective is to \textbf{maximize the yield of unique, oracle-validated counterfactuals} after canonicalization:
\begin{equation}
    \begin{aligned}
    &\max_{\mathcal{S}\subseteq \mathcal{A}(x_{\text{query}})} \ \left| \text{Unique}(\mathcal{S}) \right| \\
    &\text{s.t.}\quad \forall x' \in \mathcal{S},\ f(x')=1,\ \text{and}\ \text{cost}(\mathcal{S}) \le B_{\mathrm{LLM}}.
    \end{aligned}
\end{equation}
Here, $\text{Unique}(\mathcal{S})=\{\phi(x')\mid x'\in\mathcal{S}\}$ denotes the set of distinct canonical keys induced by a canonicalization function $\phi(\cdot)$ (defined in Sec.~\ref{sec:canonicalization}).
We define $\text{cost}(\mathcal{S})$ as the number of LLM calls \revb{used to generate} the elements of $\mathcal{S}$.%

\begin{revblock}
\pruningrev{We budget LLM calls because candidate generation typically dominates search cost and is directly controllable. Oracle evaluations $N_{\mathrm{oracle}}$ are reported separately for comparison with oracle-budgeted baselines (Sec.~\ref{sec:experiments}).}
\end{revblock}

\begin{revblock}
\pruningrev{Secondary objectives favor small changes (\textbf{proximity}), changes to few features (\textbf{sparsity}), and diversity within the canonicalized returned set (\textbf{novelty}). These preferences guide tree search through the shaped reward (Sec.~\ref{sec:reward}); the full objective is given in Eq.~\eqref{eq:multi-objective-tuple}.}
\end{revblock}

We restrict interventions to \emph{actionable} features and treat the target label, identifiers, and immutable attributes as \emph{non-actionable}.

\subsection{Preliminaries: MCTS}
\label{sec:mcts_preliminaries}
 MCTS~\citep{browne2012survey, kocsis2006bandit} is an iterative search framework that\rev{, starting from a root node,} builds a search tree through four phases: selection, expansion, simulation, and backpropagation.
 \pruningrev{In selection, the algorithm repeatedly chooses the child with the highest UCT score, $\mathrm{UCT}(n)=W(n)/N(n)+c\sqrt{\log N(\mathrm{par}(n))/N(n)}$.}
 Here, $W(n)$ is the cumulative reward \revb{for} node $n$, $N(n)$ is the number of visits to $n$, $\mathrm{par}(n)$ denotes the parent of $n$, and $c>0$ is an exploration constant controlling the exploration--exploitation trade-off. In our experiments, we set $c=1.414~(=\sqrt{2})$. For unvisited nodes ($N(n)=0$), we set $\mathrm{UCT}(n)=+\infty$ so that each node is selected at least once.
 Selection continues until a leaf node is reached, after which a new child node is added (expansion).
 The new node is then evaluated to obtain a reward $r$ (simulation), computed by a task-specific reward function.
 In the backpropagation phase, for each node $n$ on the path from the leaf to the root, we update $N(n)\leftarrow N(n)+1$ and $W(n)\leftarrow W(n)+r$.
 These phases are repeated until a predefined stopping criterion is met.

 \subsection{Proposed Method: ReCo-MCTS}
\label{sec:comp_mcts}
\pruningrev{\textbf{ReCo-MCTS} combines three components within MCTS:
(1) \textbf{Multi-candidate Expansion} to obtain $K$ proposals per LLM call,
(2) \textbf{Prompt-as-Memory} to supply records of previous proposals and outcomes, and
(3) \textbf{Oracle Evaluation and Reward Feedback} to assess constraint-admissible candidates and update tree statistics.}

\begin{figure}[t]
  \centering
\includegraphics[width=0.6\linewidth]{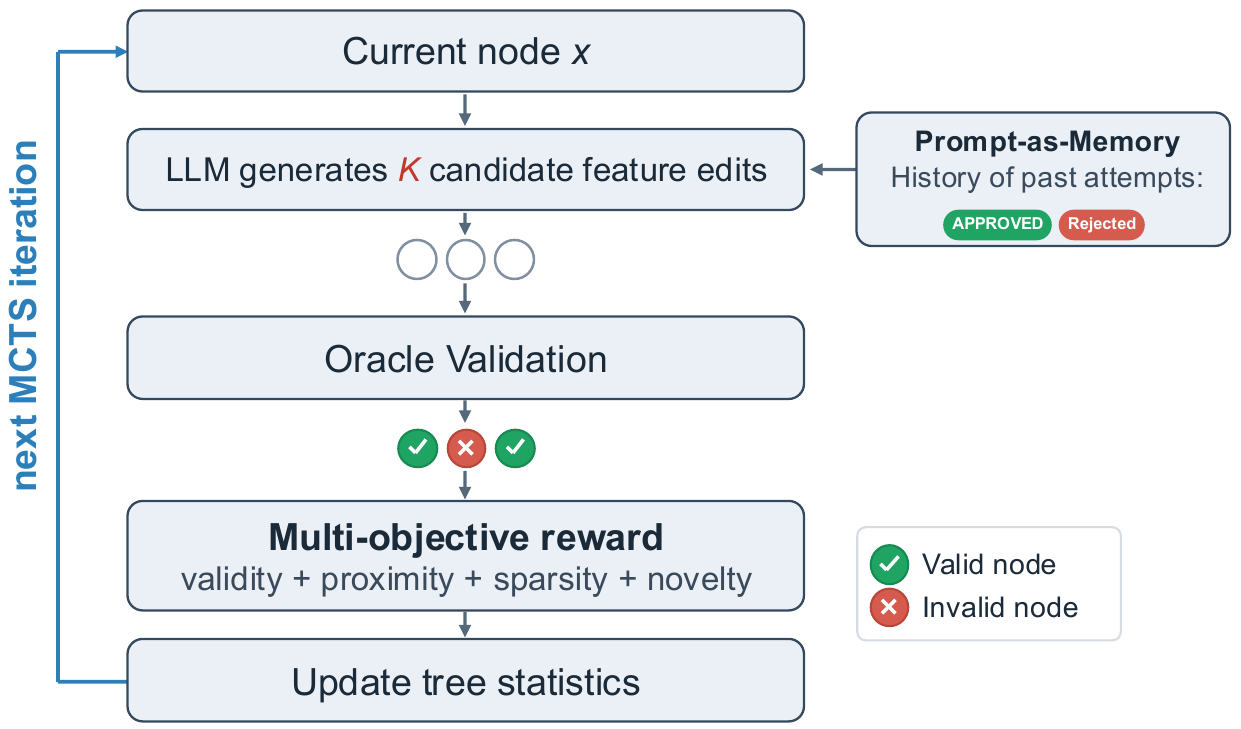}%
\vspace{-0.2cm}
  \caption{\rev{\textbf{Illustration of one iteration of ReCo-MCTS.}
}
}\label{fig:comp_mcts_pipeline}
  \Description{One ReCo-MCTS iteration: the current state and proposal history condition LLM generation of K candidates; oracle outcomes and the multi-objective reward guide updates to tree statistics.}
\vspace{-0.5cm}

\end{figure}

\looseness=-1 \pruningrev{ReCo-MCTS follows the four-phase MCTS loop (Selection, Expansion, Simulation, Backpropagation), using LLM-based candidate generation in Expansion and oracle-based candidate evaluation in place of a separate rollout in Simulation.}
Figure~\ref{fig:comp_mcts_pipeline} illustrates one iteration of the ReCo-MCTS loop.
\pruningrev{For generated candidates, oracle evaluations are bounded by $N_{\mathrm{oracle}}\le K\,B_{\mathrm{LLM}}$, excluding the initial input prediction. Constraint failures and exact prediction-cache reuse can reduce this number.}

\looseness=-1 \pruningrev{Each node $n$ in the search tree stores: (i) \textbf{Candidate state}: a counterfactual candidate $x(n)\in\mathcal{X}$; (ii) \textbf{Statistics}: visit count $N$ and cumulative value $W$ for UCT score calculation; and (iii) \textbf{Oracle outcome}: the oracle result for $x(n)$. Proposal records and their outcomes are retained for prompt construction.}
\pruningrev{For the $k$th proposal from node $n$, we write $n_k$ for the child created if the proposal passes constraint checking. Admissible candidates become children after oracle evaluation, including those that do not achieve the desired prediction. Reusing an exact cached prediction does not remove a candidate or prevent its child creation and reward update.}
Note that when the node is clear from context, we write $x$ instead of $x(n)$ for simplicity.

\subsubsection{Selection Phase}
\label{sec:mcts_selection}
\pruningrev{Selection follows UCT (Sec.~\ref{sec:mcts_preliminaries}) from the root, stopping at a childless node or depth $D_{\max}-1$; ties are broken uniformly at random. With root depth zero and $D_{\max}=5$, depth-four nodes can be expanded repeatedly, while their depth-five children are evaluated but never expanded.}

\subsubsection{Expansion Phase (Multi-Candidate)}
\pruningrev{At the selected node $n$, one LLM call produces $K$ proposals, $\{y_1, \dots, y_K\} \sim P_{\mathrm{LLM}}(\cdot \mid x(n), H_{\text{ctx}}(x(n)))$,}
where $H_{\text{ctx}}(x(n))$ is \revb{the prompt context: a compact summary} of the exploration so far (constructed as described in \hyperref[sec:prompt_memory]{Prompt-as-Memory}).

\pruningrev{Each proposal $y_k$ is a JSON list of edits to at most $L$ distinct actionable features. We apply the edits simultaneously to a copy of the parent state $x(n)$ to construct $x(n_k)$. Each listed edit must change its value. We reject proposals violating the parent-relative limit $L$ or the constraints in $\mathcal{A}(x_{\text{query}})$, including the separate limit of five changed features from the original input.}

\pruningrev{For example, a proposal may update income alone or, when $L\geq2$, update income and loan amount together. Unspecified fields retain their parent values, except for declared derived fields, which are recomputed.}

The full prompt templates and output formatting constraints are provided in \fv{Appendix~\ref{app:prompts}}{the full version}.

\phantomsection
\label{sec:prompt_memory}
\paragraph{Prompt-as-Memory.} 
\pruningrev{The prompt context $H_{\text{ctx}}(x(n))$ contains recent proposal records from all search branches, paired with status tags indicating oracle approval, rejection, or validation failure. We include the most recent records that fit within a 1,024-token history budget and a 4,096-token total input budget, with no fixed limit on the number of records (Sec.~\ref{sec:experimental_setup}). Stored records are retained even when omitted from the prompt.}

\subsubsection{\pruningrev{Canonicalization for Evaluation}}
\label{sec:canonicalization}
\pruningrev{For reporting distinct CFs, the deterministic key $\phi(x)$ groups
states with matching categorical values and numerical values in the same
training-defined ten-quantile bins, after the frozen feature normalization.
Thus, $\operatorname{Unique}(\mathcal S)=\{\phi(x):x\in\mathcal S\}$ is the
set of reporting keys. The bin boundaries are shared by all methods and
are fixed using the oracle's training split. Their construction and the
choice of the first valid representative are specified in
Appendix~\ref{app:evaluation_details}.}

\pruningrev{This grouping is an evaluation convention, not a rule for removing
search candidates. Oracle caching uses exact normalized inputs rather than
bin keys; binning does not alter the candidate values used for oracle
prediction or reward computation. Exact-state counts and alternative bin
resolutions are reported in Appendix~\ref{app:count_sensitivity}.}

\subsubsection{Simulation Phase (Multi-Objective Reward)}
\label{sec:reward}
In the simulation phase, we evaluate each candidate node $n_k$ by the oracle $f$ and assign a reward $r(x(n_k))$.
A standard reward provides an important exploration signal \revb{but does not explicitly optimize counterfactual \emph{quality}}---in particular, actionability (proximity) and recourse simplicity (sparsity). %
\rev{Moreover, auxiliary signals can be high even when the oracle acceptance probability is low; this can mislead the tree search into exploiting ``safe but invalid'' edits.}
\begin{revblock}
To address this, we present a multi-objective shaped reward in $[0,1]$ that prioritizes oracle validity while \emph{guiding the tree search} toward promising candidates via a soft gate $g(p)$.
Let $p(x(n_k))\in[0,1]$ denote the oracle's estimated acceptance probability for the candidate $x(n_k)$; if the oracle does not provide such a probability estimate, we set $p(x(n_k))=f(x(n_k))$.
The reward is defined as follows:
\end{revblock}
\begin{equation}
\begin{aligned}
r(x(n_k)) &= g(p(x(n_k))) \times \Bigl( w_1 p(x(n_k)) + w_2 s_{\text{prox}}(x(n_k),x_{\text{query}}) \\
&\hspace{4.1em} + w_3 s_{\text{spar}}(x(n_k),x_{\text{query}}) + w_4 s_{\text{nov}}(x(n_k)) \Bigr).
\end{aligned}
\label{eq:multi-objective-reward}
\end{equation}
Here, the raw weights $(w_1,w_2,w_3,w_4)$ are internally normalized to sum to 1 (i.e., $w_1 + w_2 + w_3 + w_4 = 1$).
\revb{$g(p)=1/(1+\exp(-(p-\tau)/s))$ is a soft gate that suppresses low-probability candidates, where $\tau \in [0,1]$ and $s > 0$ are hyperparameters (default $\tau=0.5$, $s=0.1$).}

\revb{The proximity score $s_{\text{prox}}(x',x_{\text{query}})=1/(1+d(x',x_{\text{query}}))$ measures how close the candidate $x'$ is to $x_{\text{query}}$.
Here, $d(x',x)=\alpha\, d_{\text{num}}(x',x) + (1-\alpha)\, d_{\text{cat}}(x',x)$ is the mixed distance combining a MAD-normalized $L_2$ distance $d_{\text{num}}$ for numerical features and a mean Hamming distance $d_{\text{cat}}$ for categorical features, where $\alpha\in[0,1]$ is a mixing weight ($\alpha=0.5$ in our experiments).}
\revb{For numerical features $\mathcal{I}_{\text{num}}$, $d_{\text{num}}(x',x)=\sqrt{\tfrac{1}{\lvert\mathcal{I}_{\text{num}}\rvert}\sum_{i\in\mathcal{I}_{\text{num}}}\bigl((x'_i-x_i)/\mathrm{MAD}_i\bigr)^2}$ is a dimension-normalized (RMS) variant based on MAD-normalized per-feature differences.
Here, $\mathrm{MAD}_i=\mathrm{median}(|x_i-\mathrm{median}(x_i)|)$ is the median absolute deviation of feature $i$, with medians taken over the training data $\mathcal{D}_{\text{train}}$ of the oracle classifier.}

\revb{For categorical features $\mathcal{I}_{\text{cat}}$, the mean Hamming distance is $d_{\text{cat}}(x',x)=\tfrac{1}{|\mathcal{I}_{\text{cat}}|}\sum_{i\in\mathcal{I}_{\text{cat}}}\mathbb{I}[x'_i\neq x_i]$, where $\mathbb{I}[\cdot]$ is the indicator function.}

\revb{The sparsity score $s_{\text{spar}}(x',x_{\text{query}})=1/(1+|\{j: x'_j \neq x_{\text{query},j}\}|)$ is the inverse count of features changed in the candidate $x'$ relative to $x_{\text{query}}$.}

\pruningrev{For each LLM call, let $\mathcal{S}_{\text{valid}}$ be a snapshot of the previously found valid states, deduplicated by exact normalized values. All $K$ candidates use this same snapshot; new valid states are added only after the call's rewards are computed. No bin aggregation is used for this reference set. The novelty reward is}
\begin{equation}
s_{\text{nov}}(x')=
\begin{cases}
1, & \text{if } \mathcal{S}_{\text{valid}}=\emptyset,\\
1-\frac{1}{1+\min_{\tilde{x}\in \mathcal{S}_{\text{valid}}} d(x',\tilde{x})}, & \text{otherwise}.
\end{cases}
\end{equation}

\label{sec:backpropagation}
\pruningrev{Finally (backpropagation), each constraint-admissible candidate supplies a reward update from its child node $n_k$ to the root: every node on that path receives $N \leftarrow N+1$ and $W \leftarrow W+r(x(n_k))$. A missing, unparsable, or constraint-violating proposal slot supplies a zero-reward update starting at the selected parent.}

\subsubsection{Algorithm Summary}
\label{sec:algorithm}
\pruningrev{The set-level objective combines CF count, proximity, sparsity, and final-set novelty:}
\pruningrev{Let $\mathcal{S}_{\text{uniq}}$ contain the first valid representative per reporting key. Final-set Novelty $D(\mathcal{S}_{\text{uniq}})$ is the mean nearest-neighbor distance (Appendix~\ref{app:evaluation_details}), distinct from the history-dependent novelty reward:}
\begin{equation}
\label{eq:multi-objective-tuple}
\begin{aligned}
&\max_{\mathcal{S}\subseteq \mathcal{A}(x_{\text{query}})} \Biggl( |\mathcal{S}_{\text{uniq}}|,\ 
\frac{1}{|\mathcal{S}_{\text{uniq}}|}\sum_{x'\in\mathcal{S}_{\text{uniq}}} s_{\text{prox}}(x',x_{\text{query}}),\\
&\hspace{1.95em}\frac{1}{|\mathcal{S}_{\text{uniq}}|}\sum_{x'\in\mathcal{S}_{\text{uniq}}} s_{\text{spar}}(x',x_{\text{query}}),\ 
\pruningrev{D(\mathcal{S}_{\text{uniq}})} \Biggr)\\
&\text{s.t.} \quad \forall x' \in \mathcal{S},\ f(x')=1,\ \text{and}\ \text{cost}(\mathcal{S}) \le B_{\mathrm{LLM}}.
\end{aligned}
\end{equation}
\noindent\pruningrev{For this objective only, quality terms are set to zero for an empty set; $D$ is also zero for a singleton. Reported empty-set quality remains undefined.}

A complete pseudocode of ReCo-MCTS is provided in \fv{Appendix~\ref{app:comp_mcts_pseudocode} (Algorithm~\ref{alg:compression_mcts})}{the full version}.

\section{Experiments}
\label{sec:experiments}

We examine two choices in ReCo-MCTS (Recourse Monte Carlo Tree Search): the number of candidates proposed
per call ($K$) and the maximum number of features changed in each proposal
($L$). We then compare the resulting counterfactual (CF) sets with those
returned by existing methods. Our primary outcome is the number of valid CFs;
proximity and sparsity characterize the changes required to obtain them.

\subsection{Experimental Setup}
\label{sec:experimental_setup}

\paragraph{Data and common constraints.}
We use Loan~\citep{loan_dataset}, Adult~\citep{kohavi1996scaling},
Credit~\citep{yeh2009comparisons}, and HELOC~\citep{heloc_kaggle}, with the
same 100 evaluation inputs per dataset for every condition (400 in total).
The previously fixed, class-weighted classifiers serve as the oracles.
A valid CF must satisfy the common feature constraints and receive the
desired oracle label. Constraints include feature types, permitted ranges
and categories, immutable features, and at most five changed features
relative to the original input. We retain the frozen preprocessing,
distance scales, and evaluation inputs throughout the comparisons.

\paragraph{ReCo-MCTS settings.}
Each run uses 30 LLM calls, UCT with exploration constant $\sqrt{2}$, maximum
tree depth five, and \experimentrev{fixed reward weights of
$(1.0,0.5,0.5,0.2)$ before normalization
(Eq.~\ref{eq:multi-objective-reward}).} Compression-based pruning is disabled.
The four settings are $(K,L)\in\{(1,3),(5,1),(5,3),(5,5)\}$.
Here, $L$ limits edits from the \emph{parent} to a proposed child; the
five-feature constraint relative to the \emph{original input} applies
separately. Recent candidate records from all branches are supplied within
a 1,024-token history budget and a 4,096-token total input budget, with no
fixed cap on the number of records. All four settings use this same history
rule and reward. The reference setting is $K=5,L=3$.
\experimentrev{We use GLM-5.3 (the deployed Inferact NVFP4 configuration)
for the main experiments and report additional results with
Qwen3.6-35B-A3B (BF16) in Appendix~\ref{app:qwen_results}.}
Each condition has one run per input with
generation seed 42. Reward details are in Appendix~\ref{app:experiment_settings}.

\paragraph{Evaluated sets and metrics.}
We evaluate each method's final output. For ReCo-MCTS this is the complete archive
of valid candidates at termination, without a final-node or top-$k$ filter.
For the baselines it is the union of valid CFs in completed native returns
across restarts within the budget; internal candidates that were never
returned are excluded. Repeated identical CFs count once. Our primary
count also groups states with matching categorical values and numerical
values in the same training-defined ten-quantile bins, retaining the first
valid state in each group. This count measures distinct alternatives at
the chosen resolution, rather than every numerical variation.
Exact-state counts and alternative bin resolutions are reported in
Appendix~\ref{app:count_sensitivity}.

\experimentrev{For input $x$, we report
$\mathrm{CF}(x)=|C(x)|$,
$\mathrm{P}(x)=\frac{1}{|C(x)|}\sum_{c\in C(x)}\frac{1}{1+d(x,c)}$,
and $\mathrm{S}(x)=\frac{1}{|C(x)|}\sum_{c\in C(x)}s(x,c)$,
where $C(x)$ is its representative CF set,}
$d$ is the frozen mixed numerical/categorical distance and $s$ counts
changed features. Higher proximity P and lower sparsity S are preferable.
CF counts include all 100 inputs, including failures. P and S first average
over CFs for each successful input, then equally over successful inputs;
$n_+$ reports this latter denominator.
\experimentrev{Bold CF counts mark the largest value per dataset.}
Overall values weight the four
dataset means equally. Paired quality comparisons use only inputs successful
under both conditions. Appendix~\ref{app:evaluation_details} specifies the
distance, aggregation, and supplementary metrics.

\paragraph{Scope of statistical analysis.}
Differences use 95\% paired bootstrap intervals with 10,000 resamples within
each dataset. These describe input-level uncertainty conditional on the
single generation seed. \experimentrev{The decisions to emphasize final
returned sets and select the model for the main text were made after
observing results.} We therefore treat the analyses as exploratory, rather than as
independent confirmatory tests. Appendix~\ref{app:returned_statistics}
\experimentrev{reports the corresponding statistics} and the multiple-comparison rule.

\subsection{Analysis of Candidate Generation Settings}
\label{sec:candidate_settings}

\begin{table}[t]
\centering
\begin{minipage}[t]{0.42\linewidth}
\vspace{0pt}
\centering
\fontsize{8}{9}\selectfont
\setlength{\tabcolsep}{3pt}
\caption{\experimentrev{Effects of Candidates per LLM Call ($K$) and the Edit Limit ($L$).}}
\label{tab:glm_settings}
\begin{fittoiclr}
\begin{tabular}{lrrrrrr}
\toprule
Dataset & $K$ & $L$ & CF $\uparrow$ & P $\uparrow$ & S $\downarrow$ & $n_+$ \\
\midrule
Loan & 1 & 3 & 8.61 & 0.677 & 2.55 & 100 \\[0.25\normalbaselineskip]
 & 5 & 1 & 69.73 & 0.661 & 2.53 & 100 \\[0.25\normalbaselineskip]
 & 5 & 3 & \textbf{76.61} & 0.692 & 3.70 & 100 \\[0.25\normalbaselineskip]
 & 5 & 5 & 72.12 & 0.694 & 3.71 & 100 \\[0.25\normalbaselineskip]
\midrule
Adult & 1 & 3 & 6.20 & 0.017 & 4.36 & 100 \\[0.25\normalbaselineskip]
 & 5 & 1 & 31.39 & 0.125 & 2.63 & 100 \\[0.25\normalbaselineskip]
 & 5 & 3 & \textbf{51.36} & 0.055 & 3.93 & 100 \\[0.25\normalbaselineskip]
 & 5 & 5 & 50.63 & 0.056 & 3.95 & 100 \\[0.25\normalbaselineskip]
\midrule
Credit & 1 & 3 & 10.33 & 0.600 & 4.49 & 96 \\[0.25\normalbaselineskip]
 & 5 & 1 & 61.85 & 0.671 & 2.98 & 91 \\[0.25\normalbaselineskip]
 & 5 & 3 & \textbf{99.92} & 0.620 & 4.52 & 99 \\[0.25\normalbaselineskip]
 & 5 & 5 & 96.14 & 0.624 & 4.50 & 99 \\[0.25\normalbaselineskip]
\midrule
HELOC & 1 & 3 & 4.64 & 0.350 & 4.77 & 91 \\[0.25\normalbaselineskip]
 & 5 & 1 & 47.03 & 0.440 & 2.91 & 96 \\[0.25\normalbaselineskip]
 & 5 & 3 & \textbf{72.91} & 0.373 & 4.74 & 96 \\[0.25\normalbaselineskip]
 & 5 & 5 & 67.30 & 0.362 & 4.77 & 100 \\[0.25\normalbaselineskip]
\bottomrule
\end{tabular}
\end{fittoiclr}
\end{minipage}
\hfill
\begin{minipage}[t]{0.56\linewidth}
\vspace{0pt}
\centering
\fontsize{8}{9}\selectfont
\setlength{\tabcolsep}{3pt}
\caption{\experimentrev{Comparison with Existing Counterfactual Methods.}}
\label{tab:glm_baselines}
\begin{fittoiclr}
\begin{tabular}{llrrrr}
\toprule
Dataset & Method & CF $\uparrow$ & P $\uparrow$ & S $\downarrow$ & $n_+$ \\
\midrule
Loan & ReCo-MCTS ($L=3$) & \textbf{76.61} & 0.692 & 3.70 & 100 \\
 & ReCo-MCTS ($L=1$) & 69.73 & 0.661 & 2.53 & 100 \\
 & DiCE & 24.57 & 0.701 & 1.77 & 100 \\
 & Growing Spheres & 0.00 & -- & -- & 0 \\
 & \experimentrev{C-CHVAE} & 0.00 & -- & -- & 0 \\
\midrule
Adult & ReCo-MCTS ($L=3$) & \textbf{51.36} & 0.055 & 3.93 & 100 \\
 & ReCo-MCTS ($L=1$) & 31.39 & 0.125 & 2.63 & 100 \\
 & DiCE & 16.65 & 0.028 & 1.96 & 100 \\
 & Growing Spheres & 0.02 & 0.243 & 1.00 & 2 \\
 & \experimentrev{C-CHVAE} & 0.04 & 0.140 & 3.00 & 4 \\
\midrule
Credit & ReCo-MCTS ($L=3$) & \textbf{99.92} & 0.620 & 4.52 & 99 \\
 & ReCo-MCTS ($L=1$) & 61.85 & 0.671 & 2.98 & 91 \\
 & DiCE & 19.05 & 0.249 & 2.30 & 98 \\
 & Growing Spheres & 0.02 & 0.999 & 1.50 & 2 \\
 & \experimentrev{C-CHVAE} & 0.00 & -- & -- & 0 \\
\midrule
HELOC & ReCo-MCTS ($L=3$) & \textbf{72.91} & 0.373 & 4.74 & 96 \\
 & ReCo-MCTS ($L=1$) & 47.03 & 0.440 & 2.91 & 96 \\
 & DiCE & 23.88 & 0.397 & 2.15 & 100 \\
 & Growing Spheres & 0.09 & 0.966 & 1.38 & 8 \\
 & \experimentrev{C-CHVAE} & 0.00 & -- & -- & 0 \\
\bottomrule
\end{tabular}
\end{fittoiclr}
\end{minipage}

\end{table}

\paragraph{Number of candidates per call.}
With $L=3$, increasing $K$ from one to five raises the mean CF count on all
four datasets (Table~\ref{tab:glm_settings}). The equal-dataset mean rises
from 7.45 to 75.20, a difference of 67.75 CFs
(95\% CI $[65.28,70.21]$). Successful inputs increase from 387 to 395 out
of 400. This establishes a quantity benefit from multiple-candidate
expansion under a fixed number of LLM calls. It also increases the maximum
number of requested candidates from 30 to 150 and changes tree branching;
the comparison does not hold candidate count or token expenditure constant.
Mean total tokens rise from 70.2k to 87.9k and recorded search time from
58.5 to 232.2 seconds per input. The result concerns the combined effect
of this expansion setting, rather than isolating UCT or history.

\paragraph{Maximum edits per proposal.}
With $K=5$, $L=3$ gives the largest mean CF count in each dataset.
Its overall count is 75.20, compared with 52.50 for $L=1$ and 71.55
for $L=5$. The $L=3$ minus $L=1$ difference is 22.70
($[20.15,25.31]$); increasing the limit from three to five does not
increase the overall count (difference $-3.65$, $[-5.71,-1.57]$).
This is an observed comparison among the tested settings, not evidence
that $L=3$ is universally optimal. In particular, $L=5$ succeeds on
399 inputs, versus 395 for $L=3$, despite returning fewer CFs on average.

The additional CFs require different interventions. $L=1$ has fewer
changed features than $L=3$ on all four datasets, with overall S of
2.76 versus 4.22. It also has higher P on Adult, Credit, and HELOC;
Loan is the exception, where $L=3$ is closer. Thus, allowing several
edits per proposal increases the range of alternatives in these runs,
while single-feature proposals usually produce smaller interventions.
The sparsity comparison retains the same direction on common-success
inputs. Qwen also favors $K=5$ for CF count and has its largest count
at $L=3$ on each dataset, but its success coverage and proximity trade-offs
differ (Appendix~\ref{app:qwen_results}).

\subsection{Comparison with Existing Methods}
\label{sec:comparison_existing}

\paragraph{Methods and budgets.}
We compare \experimentrev{ReCo-MCTS with $K=5$ and $L\in\{1,3\}$}
with DiCE~\citep{mothilal2020dice}, Growing
Spheres~\citep{laugel2018comparison}, and C-CHVAE~\citep{pawelczyk2020learning}.
DiCE uses the official random implementation; Growing Spheres uses the
authors' search with a mixed-feature adapter; C-CHVAE uses the CARLA
PyTorch implementation with a mixed-feature adapter. These implementation
choices and information access are detailed in
Appendix~\ref{app:baseline_settings}.
Baselines receive limits of $B\in\{150,1{,}000,10{,}000\}$
\experimentrev{oracle evaluations}, including evaluations during postprocessing.
\experimentrev{We count one oracle evaluation per candidate evaluated by
the classifier. Each candidate in a batch is counted separately;
cache hits incur no new evaluation.}
\experimentrev{Table~\ref{tab:glm_baselines} shows the largest tested budget,
$B=10{,}000$, because DiCE returns no CFs before interruption at
$B=150$ or $1{,}000$, leaving P and S undefined. All three budgets were
planned; the choice to feature the largest in the main table was made
after observing results. All budgets are retained in
Appendix~\ref{app:baseline_budgets}.} DiCE requests five
CFs per restart, so its accumulated output is not capped at five.
The ReCo-MCTS limit of 30 LLM calls and the baseline oracle limits are different
resource constraints. We report both resource use and observed time,
without interpreting them as equal-compute budgets.
\experimentrev{A dash in Table~\ref{tab:glm_baselines} indicates that
no valid CF was returned.}

\paragraph{CF count and coverage.}
\experimentrev{At $L=3$, ReCo-MCTS returns more CFs than the three
implementations in Table~\ref{tab:glm_baselines} on every dataset.}
Against DiCE at $B=10{,}000$, the mean is 75.20
versus 21.04, a difference of 54.16 ($[51.76,56.51]$).
Both cover almost all inputs: 395/400 for ReCo-MCTS and 398/400 for DiCE.
The count advantage is retained with exact-state counts and both alternative
bin resolutions. ReCo-MCTS with $L=1$ also exceeds DiCE in CF count on each
dataset, although its coverage is lower (387/400).
Growing Spheres and C-CHVAE return valid CFs for only 12 and four inputs,
respectively, at $B=10{,}000$ under the tested adapters and constraints.
Their quality averages therefore describe very small subsets.

\paragraph{Quality and cost.}
DiCE requires fewer changed features than either ReCo-MCTS setting on every
dataset. For $L=3$, its S ranges from 1.77 to 2.30, compared with
3.70 to 4.74 for ReCo-MCTS. Proximity is dataset-dependent: ReCo-MCTS has higher P
on Adult and Credit, whereas DiCE has higher P on Loan and HELOC.
With $L=1$, ReCo-MCTS improves proximity on HELOC as well, but still changes
more features than DiCE. These directions persist in common-success
comparisons; the small Loan proximity difference for $L=3$ has an
interval including zero (Appendix~\ref{app:returned_statistics}).
DiCE also has larger within-set Novelty D on all four datasets, while
ReCo-MCTS has smaller distance to the training reference (NN5); these
supplementary measures describe different aspects of the returned sets
(Appendix~\ref{app:supplementary_quality}).

ReCo-MCTS with $L=3$ uses a mean of 126.6 actual oracle evaluations, compared
with 10,000 for DiCE, but additionally consumes 30 LLM calls and 87.9k
tokens. Its recorded search time is 232.2 seconds per input, versus
1.03 seconds for DiCE in their respective execution environments
(Appendix~\ref{app:search_costs}). Hence the CF-count advantage comes with
substantially greater LLM and time costs; it is not a general efficiency
or quality advantage.

\section{Conclusions and Future Work}
\begin{revblock}
\looseness=-1 We presented Recourse Monte Carlo Tree Search (ReCo-MCTS), an agentic tree-search method for generating \emph{diverse} counterfactual recourse options under a fixed LLM-call budget.
Our contributions are summarized as follows: (i) \textbf{Problem formulation:} We formulated counterfactual recourse generation as a fixed-budget search problem that aims to maximize the yield of \emph{unique, oracle-validated} counterfactuals while maintaining favorable quantity--quality trade-offs, rather than converging to a single best counterfactual (Sec.~\ref{sec:problem_formulation}); 
(ii) \textbf{Search framework:} We proposed ReCo-MCTS, a budget-aware agentic tree-search method that combines LLM-based multi-candidate generation, constraint checking, black-box oracle evaluation, and UCT-guided search to accumulate valid counterfactuals across explored branches (Sec.~\ref{sec:comp_mcts});
(iii) \textbf{Empirical findings:} \experimentrev{On four tabular datasets, multiple-candidate generation increased CF count, while the edit limit changed the quantity--quality trade-off. ReCo-MCTS returned more CFs than the tested DiCE, Growing Spheres, and C-CHVAE implementations; DiCE changed fewer features and had lower recorded search time. These results do not establish an equal-compute advantage (Sec.~\ref{sec:experiments}).}

\revb{Future work includes extending the approach beyond low-dimensional tabular domains, incorporating richer feasibility constraints, and studying theoretical guarantees and human-in-the-loop evaluation.}
\end{revblock}

\section*{AI Use Statement}
We used large language models to assist with revising the manuscript and
implementing the proposed method. The authors take responsibility for the
final content of this work, including the text, implementation, and reported
results.

\bibliographystyle{iclr2027_conference}
\bibliography{references}

@inproceedings{yao2023tree,
  title        = {Tree of Thoughts: Deliberate Problem Solving with Large Language Models},
  author       = {Yao, Shunyu and Yu, Dian and Zhao, Jeffrey and Shafran, Izhak and Griffiths, Thomas L. and Cao, Yuan and Narasimhan, Karthik},
  booktitle    = {Advances in Neural Information Processing Systems},
  year         = {2023},
  volume       = {36},
  doi          = {10.52202/075280-0517}
}

@misc{loan_dataset,
  title={Loan-Approval-Prediction-Dataset},
  author={{Kaggle}},
  howpublished={\url{https://www.kaggle.com/datasets/architsharma01/loan-approval-prediction-dataset}},
  year={n.d.},
  note={Accessed: 2026-09-26}
}

@inproceedings{kohavi1996scaling,
  author       = {Kohavi, Ron},
  title        = {Scaling Up the Accuracy of {Naive-Bayes} Classifiers: A Decision-Tree Hybrid},
  booktitle    = {Proceedings of the Second International Conference on Knowledge Discovery and Data Mining (KDD)},
  year         = {1996},
  note         = {Dataset used: UCI Adult (Census Income) dataset},
  pages        = {202--207},
  publisher    = {AAAI Press}
}

@article{yeh2009comparisons,
  author       = {Yeh, I-Cheng and Lien, Che-hui},
  title        = {The Comparisons of Data Mining Techniques for the Predictive Accuracy of Probability of Default of Credit Card Clients},
  journal      = {Expert Systems with Applications},
  year         = {2009},
  volume       = {36},
  number       = {2},
  pages        = {2473--2480},
  doi          = {10.1016/j.eswa.2007.12.020}
}

@misc{heloc_kaggle,
  title        = {Home Equity Line of Credit ({HELOC})},
  author       = {{Kaggle}},
  year         = {2024},
  howpublished = {\url{https://www.kaggle.com/datasets/averkiyoliabev/home-equity-line-of-creditheloc}},
  note         = {Accessed: 2026-01-24}
}

@article{wachter2018counterfactual,
  title        = {Counterfactual Explanations Without Opening the Black Box: Automated Decisions and the {GDPR}},
  author       = {Wachter, Sandra and Mittelstadt, Brent and Russell, Chris},
  journal      = {Harvard Journal of Law \& Technology},
  year         = {2018},
  volume       = {31},
  number       = {2}
}

@inproceedings{ustun2019actionable,
  title        = {Actionable Recourse in Linear Classification},
  author       = {Ustun, Berk and Spangher, Alexander and Liu, Yang},
  booktitle    = {Proceedings of the Conference on Fairness, Accountability, and Transparency (FAT*)},
  year         = {2019},
  doi          = {10.1145/3287560.3287566}
}

@inproceedings{mothilal2020dice,
  title        = {Explaining Machine Learning Classifiers through Diverse Counterfactual Explanations},
  author       = {Mothilal, Ramaravind K. and Sharma, Amit and Tan, Chenhao},
  booktitle    = {Proceedings of the 2020 Conference on Fairness, Accountability, and Transparency},
  year         = {2020},
  pages        = {607--617},
  doi          = {10.1145/3351095.3372850}
}

@inproceedings{pawelczyk2020learning,
  title        = {Learning Model-Agnostic Counterfactual Explanations for Tabular Data},
  author       = {Pawelczyk, Martin and Broelemann, Klaus and Kasneci, Gjergji},
  booktitle    = {Proceedings of The Web Conference 2020},
  year         = {2020},
  pages        = {3126--3132},
  publisher    = {Association for Computing Machinery}
}

@inproceedings{pawelczyk22a,
  title        = {Exploring Counterfactual Explanations Through the Lens of Adversarial Examples: A Theoretical and Empirical Analysis},
  author       = {Pawelczyk, Martin and Agarwal, Chirag and Joshi, Shalmali and Upadhyay, Sohini and Lakkaraju, Himabindu},
  booktitle    = {Proceedings of the 25th International Conference on Artificial Intelligence and Statistics (AISTATS)},
  year         = {2022},
  series       = {Proceedings of Machine Learning Research},
  volume       = {151},
  pages        = {4574--4594},
  publisher    = {PMLR},
  url          = {https://proceedings.mlr.press/v151/pawelczyk22a.html}
}

@inproceedings{karimi2021algorithmic,
  title        = {Algorithmic Recourse: from Counterfactual Explanations to Interventions},
  author       = {Karimi, Amir-Hossein and Sch{\"o}lkopf, Bernhard and Valera, Isabel},
  booktitle    = {Proceedings of the 2021 ACM Conference on Fairness, Accountability, and Transparency (FAccT '21)},
  year         = {2021},
  doi          = {10.1145/3442188.3445899}
}

@inproceedings{karimi2020probabilistic,
  title        = {Algorithmic recourse under imperfect causal knowledge: a probabilistic approach},
  author       = {Karimi, Amir-Hossein and von K{\"u}gelgen, Julius and Sch{\"o}lkopf, Bernhard and Valera, Isabel},
  booktitle    = {Advances in Neural Information Processing Systems},
  year         = {2020},
  volume       = {33}
}

@inproceedings{dandl2020multi,
  title        = {Multi-Objective Counterfactual Explanations},
  author       = {Dandl, Susanne and Molnar, Christoph and Binder, Martin and Bischl, Bernd},
  booktitle    = {Parallel Problem Solving from Nature -- PPSN XVI},
  year         = {2020},
  series       = {Lecture Notes in Computer Science},
  volume       = {12269},
  pages        = {448--469},
  doi          = {10.1007/978-3-030-58112-1_31}
}

@inproceedings{bhattacharjee2024zeroshot,
  title        = {Zero-shot {LLM}-guided Counterfactual Generation: A Case Study on {NLP} Model Evaluation},
  author       = {Bhattacharjee, Amrita and Moraffah, Raha and Garland, Joshua and Liu, Huan},
  year         = {2024},
  booktitle    = {2024 IEEE International Conference on Big Data (BigData)},
  publisher    = {IEEE},
  note         = {Extended version: arXiv:2405.04793},
  url          = {https://arxiv.org/abs/2405.04793}
}

@inproceedings{laugel2018comparison,
  title        = {Comparison-Based Inverse Classification for Interpretability in Machine Learning},
  author       = {Laugel, Thibault and Lesot, Marie-Jeanne and Marsala, Christophe and Renard, Xavier and Detyniecki, Marcin},
  booktitle    = {Information Processing and Management of Uncertainty in Knowledge-Based Systems. Theory and Foundations},
  year         = {2018},
  series       = {Communications in Computer and Information Science},
  volume       = {853},
  pages        = {100--111},
  doi          = {10.1007/978-3-319-91473-2_9}
}

@inproceedings{satml24,
 title = {Navigating the Structured What-If Spaces: Counterfactual Generation via Structured Diffusion},
 author = {Nishtha Madaan and Srikanta Bedathur},
 booktitle = {Proceedings of the 2nd IEEE Conference on Secure and Trustworthy Machine Learning},
 year = {2024}
}

@inproceedings{rasal2025diffusion,
  title        = {Diffusion Counterfactual Generation with Semantic Abduction},
  author       = {Rasal, Rajat R. and Kori, Avinash and De Sousa Ribeiro, Fabio and Xia, Tian and Glocker, Ben},
  booktitle    = {Proceedings of the 42nd International Conference on Machine Learning},
  year         = {2025},
  series       = {Proceedings of Machine Learning Research},
  volume       = {267},
  pages        = {51201--51228},
  publisher    = {PMLR},
  url          = {https://proceedings.mlr.press/v267/rasal25a.html}
}

@inproceedings{Jeanneret_2022_ACCV,
  author       = {Jeanneret, Guillaume and Simon, Lo{\"i}c and Jurie, Fr{\'e}d{\'e}ric},
  title        = {Diffusion Models for Counterfactual Explanations},
  booktitle    = {Proceedings of the Asian Conference on Computer Vision},
  year         = {2022},
  pages        = {858--876}
}

@article{browne2012survey,
  title        = {A Survey of {Monte Carlo} Tree Search Methods},
  author       = {Browne, Cameron B. and Powley, Edward and Whitehouse, Daniel and Lucas, Simon M. and Cowling, Peter I. and Rohlfshagen, Philipp and Tavener, Stephen and Perez, Diego and Samothrakis, Spyridon and Colton, Simon},
  journal      = {IEEE Transactions on Computational Intelligence and AI in Games},
  year         = {2012},
  volume       = {4},
  number       = {1},
  pages        = {1--43},
  doi          = {10.1109/TCIAIG.2012.2186810}
}

@inproceedings{kocsis2006bandit,
  title        = {Bandit Based {Monte-Carlo} Planning},
  author       = {Kocsis, Levente and Szepesv{\'a}ri, Csaba},
  booktitle    = {Machine Learning: ECML 2006},
  year         = {2006},
  series       = {Lecture Notes in Computer Science},
  volume       = {4212},
  pages        = {282--293},
  doi          = {10.1007/11871842_29}
}

@inproceedings{sharma2020certifai,
  author       = {Sharma, Shubham and Henderson, Jette and Ghosh, Joydeep},
  title        = {{CERTIFAI}: A Common Framework to Provide Explanations and Analyse the Fairness and Robustness of Black-box Models},
  year         = {2020},
  booktitle    = {Proceedings of the AAAI/ACM Conference on AI, Ethics, and Society}
}

@misc{dice_code,
  key          = {DiCE},
  title        = {{DiCE}: Diverse Counterfactual Explanations},
  howpublished = {\url{https://github.com/interpretml/DiCE}},
  year         = {2019},
  note         = {Accessed: 2026-02-23}

}

@inproceedings{verma2022amortized,
  author       = {Verma, Sahil and Hines, Keegan and Dickerson, John P.},
  title        = {Amortized Generation of Sequential Algorithmic Recourses for Black-Box Models},
  booktitle    = {Proceedings of the AAAI Conference on Artificial Intelligence},
  volume       = {36},
  note         = {Issue 8},
  pages        = {8512--8519},
  year         = {2022},
  doi          = {10.1609/aaai.v36i8.20828}
}

@inproceedings{vo2023feature,
  author    = {Vo, Vy and Le, Trung and Nguyen, Van and Zhao, He and Bonilla, Edwin V. and Haffari, Gholamreza and Phung, Dinh},
  title     = {Feature-Based Learning for Diverse and Privacy-Preserving Counterfactual Explanations},
  booktitle = {Proceedings of the 29th ACM SIGKDD Conference on Knowledge Discovery and Data Mining},
  pages     = {2211--2222},
  year      = {2023}
}

@inproceedings{russell2019efficient,
  author       = {Russell, Chris},
  title        = {Efficient Search for Diverse Coherent Explanations},
  booktitle    = {Proceedings of the Conference on Fairness, Accountability, and Transparency},
  pages        = {20--28},
  year         = {2019},
  doi          = {10.1145/3287560.3287569}
}

\iffullversion
\clearpage
\appendix
\section{Related Work Details}
\label{app:related_work_details}

This appendix contains the detailed related-work discussion from Sec.~2.

The following subsections review representative lines of work and discuss their limitations with respect to our setting: counterfactual recourse under a fixed inference budget, with emphasis on the LLM-agentic setting and connections to oracle-budgeted non-LLM baselines.
Throughout this paper, we use \emph{fixed inference budget} as a general abstraction of limited computational resources. Depending on the method class, this budget is instantiated either as a fixed number of LLM calls (for LLM-agentic search) or as a fixed number of oracle evaluations (for non-LLM baselines).
\relatedrev{Table~\ref{tab:comparison_methods} provides an overview of counterfactual explanation methods. DiCE, Growing Spheres, and C-CHVAE are the empirical baselines; the other listed prior methods supply context for the related-work discussion.}
\relatedrev{The table describes each method's primary objective, candidate generation and search, and returned CFs. Entries for prior methods refer to their cited formulations: DiCE optimizes a requested CF set, CERTIFAI selects a requested number by fitness, and the listed single-solution procedures can be rerun. ReCo-MCTS accumulates valid CFs across explored branches until its LLM-call budget is exhausted. The experimental variants and collection across restarts are specified in Section~\ref{sec:comparison_existing}.}

\subsection{Distance-Minimization Recourse and Its Limitations}
Early counterfactual explanation methods cast recourse as a distance-minimization problem: given an instance $x$, find the closest $x'$ that flips the model prediction while minimizing a norm- or feature-wise distance $d(x,x')$~\citep{wachter2018counterfactual,ustun2019actionable}.
\citet{laugel2018comparison} proposed Growing Spheres, an instance-based and model-agnostic approach that identifies the closest decision-boundary-crossing example using a combined proximity and sparsity criterion.
Subsequent work extends this template with additional objectives (e.g., DiCE~\citep{mothilal2020dice}) and multi-objective variants (e.g., MOC~\citep{dandl2020multi}) for promoting diversity and exposing trade-offs among competing criteria.
However, this common ``minimal perturbation'' principle also makes distance-minimizing counterfactual generation structurally close to adversarial example generation and can even become equivalent in particular hyperparameter regimes (e.g., limiting cases)~\citep{pawelczyk22a}.
As a result, purely proximity-driven objectives risk producing counterfactuals that cross the decision boundary by exploiting model vulnerabilities rather than suggesting realistic or semantically meaningful changes (the ``adversarial trap'').
Even when augmented with set-level diversity terms (e.g., DiCE), these approaches typically remain driven by input-space distance surrogates and can inherit this vulnerability when proximity dominates the objective.

\citet{sharma2020certifai} proposed CERTIFAI, a unified framework that leverages genetic-algorithm-based counterfactual generation to analyze explainability, robustness, and fairness of black-box models.
CERTIFAI introduced CERScore and a burden-based group fairness notion by measuring the distance between inputs and their corresponding counterfactuals.
While CERTIFAI connects counterfactual generation with robustness and fairness analysis, its generation procedure remains proximity-driven and does not explicitly target the fixed-budget, set-level objective emphasized in this work: maximizing the yield of \emph{unique, oracle-validated counterfactuals} (distinct counterfactuals after canonicalization) under a fixed inference budget.
Moreover, redundancy control and budget efficiency are not treated as primary objectives, which become central when the goal is to maximize yield under a strict budget.

\subsection{Causal-Structure-Based Algorithmic Recourse}
Relatedly, causal-structure-based recourse methods shift the focus from proximity-minimizing counterfactuals to \emph{interventions} in a Structural Causal Model (SCM).
\citet{karimi2021algorithmic} formulate algorithmic recourse as finding a minimal-cost intervention that accounts for downstream effects of feature changes under a specified SCM.
To relax the assumption of perfect knowledge of structural equations, they also propose a probabilistic recourse framework that reasons under uncertainty (e.g., via Bayesian inference or subpopulation-level estimation)~\citep{karimi2020probabilistic}.
While compelling when a reliable causal model is available, these approaches typically require at least a known causal graph (and often additional structural assumptions) and can be computationally demanding, making them difficult to apply in strictly black-box settings where no causal structure is specified.

\subsection{Manifold- and Generative-Model-Based Counterfactual Explanations}
\citet{pawelczyk2020learning} proposed C-CHVAE, a model-agnostic framework that generates counterfactual explanations by searching in a learned latent space that approximates the data manifold, rather than directly minimizing input-space perturbations.
By constraining the search to high-density regions of the learned distribution, C-CHVAE aims to improve plausibility and reduce unrealistic boundary-crossing artifacts.
However, C-CHVAE is primarily formulated as a single-solution search that returns one closest counterfactual per query instance.
Consequently, it does not address the problem setting where the goal is to return a large set of distinct counterfactuals under a fixed inference budget, nor does it incorporate explicit redundancy control across multiple candidate solutions.
In particular, while manifold-awareness can improve individual plausibility, it does not specify how to allocate a fixed budget across multiple candidates to avoid redundancy and return many distinct, actionable recourse options.
In contrast, our formulation casts counterfactual generation as a budgeted, diversity-seeking search problem that aims to maximize the yield of \emph{unique}, \emph{oracle-validated} counterfactuals while maintaining favorable quantity--quality trade-offs under a fixed LLM-call budget.

Recent work~\citep{Jeanneret_2022_ACCV, rasal2025diffusion} has explored the use of diffusion models for counterfactual generation, especially for image domains.
\citet{satml24} proposed Structured Counterfactual Diffusion (SCD), a diffusion-based framework for generating plausible counterfactuals in tabular data by modeling the joint distribution of structured features.
While SCD enforces plausibility and diversity through generative sampling, it does not formulate counterfactual generation as a budgeted search process, nor does it explicitly implement redundancy control or optimize the yield of unique, oracle-validated recourse options under a fixed inference budget.
Moreover, while distribution-aware generation in SCD may reduce some unrealistic boundary-crossing artifacts, label-guided sampling alone does not guarantee robustness to adversarial-trap behavior.

\subsection{\relatedrev{LLM-Based Search and Its Relation to Recourse}}
Recent work has explored the use of large language models (LLMs) as direct generators of counterfactual explanations.
\citet{bhattacharjee2024zeroshot} demonstrate that instruction-tuned LLMs can generate plausible counterfactuals in a zero-shot manner without task-specific training.
However, these approaches mainly use LLMs as flexible proposal generators and do not explicitly address systematic exploration, strict oracle validation, or budget control under a fixed LLM-call budget.
Moreover, this line of work is primarily concerned with textual counterfactuals for NLP model evaluation and does not consider feature-level interventions, oracle-validated validity checks, or actionable recourse in tabular domains.

\relatedrev{Tree-of-Thoughts considers alternative reasoning paths~\citep{yao2023tree}. In ReCo-MCTS, nodes represent tabular feature states and the output accumulates valid candidates across the explored tree.}

\relatedrev{The empirical questions concern the effects of $K$ and $L$ within ReCo-MCTS and its comparison with the three specified CF baselines.}
\pruningrev{We propose \textbf{ReCo-MCTS}, an MCTS framework that combines LLM-based multi-candidate generation, constraint checking, and oracle evaluation, using candidate rewards to guide tree selection and accumulating valid counterfactuals under a fixed LLM-call budget.}

\FloatBarrier
\section{Prompt Templates and Formatting Constraints}
\label{app:prompts}

\pruningrev{The generator receives a fixed system message and a JSON user message. The system message below is reproduced with line wrapping for readability; the user-message fields are instantiated for each selected parent. A proposal is a list of simultaneous edits, with at most $L$ changed features from the parent and at most five from the original input.}

{\ifpruningrevisionhighlight\color{pruningrevisionblue}\fi
\paragraph{System message.}
\begin{quote}
\small
\begin{verbatim}
Generate diverse counterfactual feature edits that achieve the
desired classifier label.
Return only a JSON object
{"candidates":[{"edits":[{"feature":"name","value":value}]},
...]}.
Each candidate is an independent edit list applied
simultaneously to a copy of x_parent.
Use distinct editable feature names only. Every listed edit must
change its value relative to x_parent.
Respect max_changed_features_per_transition and
max_changed_features_from_x0. Unspecified fields stay unchanged.
Never edit immutable, missing/sentinel, target, ID or derived
fields. Declared derived fields are recomputed.
Follow exact names, types, categories, units, bounds and
relations. Numeric values must be finite and on the unit grid.
Original out-of-range values may remain unchanged. Compare final
constraints, closeness and sparsity against x0.
Prefer small changes and diverse solutions; avoid duplicates.
One edit list is one candidate, not multiple candidates.
Past validation records, when present, include edits relative to
x0 and the transition edit list.
Constraint-invalid candidates have no probability. Return the
requested number of candidates.
\end{verbatim}
\end{quote}

\paragraph{User message.}
The user message is one JSON object with the following fields:
\begin{itemize}
  \item \texttt{desired\_label}: the target classifier label.
  \item \texttt{requested\_candidates}: the number $K$ requested in this call.
  \item \texttt{schema}: feature names, types, editable and immutable fields, categories, numerical units and bounds, sentinel values, derived fields, and relations. It includes \texttt{max\_changed\_features\_per\_transition} set to $L$ and \texttt{max\_changed\_features\_from\_x0} set to five.
  \item \texttt{x0}: the original rejected input.
  \item \texttt{x\_parent}: the current selected state, which may already have the target label.
  \item \texttt{memory}: the retained history records, omitted when none are included.
\end{itemize}
The message is serialized as compact JSON with sorted keys and no non-finite numerical values. The schema requires distinct editable features and rejects any candidate containing a no-op edit or another constraint violation.

\paragraph{History construction.}
History records are drawn from all search branches in chronological order.
Each record contains \texttt{call\_id}, \texttt{candidate\_id},
\texttt{edits\_from\_x0}, \texttt{label}, \texttt{p\_target}, and
\texttt{transition\_edits}. Labels are \texttt{VALID},
\texttt{ORACLE\_NEGATIVE}, or \texttt{CONSTRAINT\_INVALID}.
Constraint-invalid records additionally contain \texttt{reasons} and
\texttt{raw\_edit}, with a null probability.
Missing and unparsable proposal slots are logged but not included in prompt history.

Starting from the full stored history, the oldest records are removed from the prompt subset until its serialized history fits within 1,024 tokens and the complete messages fit within 4,096 tokens. There is no count cap, and stored history is never deleted. If the mandatory messages alone exceed the input limit, the run reports an error rather than discarding required input fields.

\subsection{Notes on Parsability and Constraints}
\label{app:prompts_notes}

The response must be a JSON object whose only key is \texttt{candidates}.
Each candidate contains only an \texttt{edits} list, and each edit contains
a \texttt{feature} name and a \texttt{value}. Every list is applied independently
to a copy of \texttt{x\_parent}; it is not a sequence of edits to earlier siblings.

Only the first $K$ candidates are processed. Missing slots and parse failures
receive zero reward. Duplicate feature names, no-op edits, and violations of
the schema or either edit limit cause the whole candidate to be rejected
before oracle evaluation. Accepted edit lists update the specified fields;
declared derived fields are then recomputed.
}

\clearpage
\section{Dataset Details}
\label{app:datasets}

This appendix provides descriptions of the four datasets.

\paragraph{Loan.}
The Loan dataset~\citep{loan_dataset} is a home-loan eligibility dataset, where each instance corresponds to a loan application described by demographic and financial attributes, and a pre-trained oracle classifies whether the applicant qualifies for a loan.
It contains 4,269 loan applications with financial and asset-related features (e.g., income, loan amount/term, credit score, and assets) and the loan status (Approved vs.\ Rejected). After excluding the target column (loan status) and identifier columns, the oracle input has 11 features (9 numerical and 2 categorical: education, self-employed).

\paragraph{Adult.}
The Adult dataset~\citep{kohavi1996scaling} is a widely used census-income dataset, where each instance corresponds to an individual described by demographic and employment attributes (e.g., education, age, sex, and occupation), and a pre-trained oracle classifies whether the income exceeds $50$K.
It contains 32,561 records with the income label ($>50$K vs.\ $\le 50$K). After excluding the target column (income) and identifier columns, the oracle input has 14 features (6 numerical and 8 categorical, of which sex, race, and native country are treated as immutable following common practice for counterfactual fairness).

\paragraph{Credit.}
The Credit dataset~\citep{yeh2009comparisons} is a credit-default dataset of credit card clients in Taiwan, spanning April--September 2005, where each instance is described by repayment status history, bill and payment amounts, and basic demographics; a pre-trained oracle classifies whether the client will default the following month.
It contains 30,000 records with the default label (default vs.\ non-default). After excluding the target column (default payment next month) and identifier columns, the oracle input has 23 features (20 numerical and 3 categorical: sex, education, marriage, of which sex is treated as immutable).

\paragraph{HELOC.}
The HELOC dataset~\citep{heloc_kaggle} is a home-equity line-of-credit risk dataset, where each instance corresponds to an anonymized credit applicant described by bureau-derived credit-history attributes, and a pre-trained oracle classifies the applicant's risk performance (Good vs.\ Bad).
It contains 10,459 records with 23 predictor features, all treated as numerical in our implementation. These features summarize account age, delinquency history, inquiry activity, utilization, and trade-balance information (e.g., external risk estimate, months since oldest trade, delinquency indicators, and revolving/installment burden ratios). After excluding the target column (risk performance), the oracle input has 23 numerical features and no categorical features.

\clearpage
\section{ReCo-MCTS Pseudocode}
\label{app:comp_mcts_pseudocode}
\pruningrev{Algorithm~\ref{alg:compression_mcts} specifies the selection, history, and reward-update rules used in the experiments. The archive $\mathcal S$ contains exact-unique valid states; reporting bins are applied only for evaluation.}

\begin{algorithm}[H]
\caption{\pruningrev{ReCo-MCTS}}
\label{alg:compression_mcts}
{\ifpruningrevisionhighlight\color{pruningrevisionblue}\fi
\begin{algorithmic}[1]
\Require Rejected input $x_0$, LLM-call budget $B_{\mathrm{LLM}}$, oracle $f,p$, generator $G$, candidates per call $K$, parent edit limit $L$, maximum depth $D_{\max}$, feature constraints and root edit limit five
\State Initialize root $n_0$ at depth zero; $\mathcal S\gets\emptyset$, history $H\gets[]$
\For{$t=1$ to $B_{\mathrm{LLM}}$}
    \State $n\gets n_0$
    \While{$n$ has children and $\operatorname{depth}(n)<D_{\max}-1$}
        \State Choose the child with maximum UCT; break ties uniformly at random
        \State $n\gets$ chosen child
    \EndWhile
    \State $\mathcal R_t\gets\mathcal S$ \Comment{Snapshot shared by this call's candidates}
    \State Build $H_{\mathrm{ctx}}$ from the most recent records in global history $H$
    \Statex \hspace{\algorithmicindent} within 1,024 history tokens and 4,096 total input tokens
    \State Generate $K$ edit lists conditioned on $x_0,x(n),L,H_{\mathrm{ctx}}$ and constraints
    \State Parse the first $K$ candidate slots; mark missing or unparsable slots
    \State $\mathcal V_t\gets\emptyset$
    \For{each candidate slot}
        \State Apply its edit list to a copy of $x(n)$ if parsable
        \If{missing, unparsable, or violating feature or edit constraints}
            \State Log the failure; append a constraint-invalid record to $H$ if applicable
            \State $\operatorname{Backpropagate}(n,0)$; \textbf{continue}
        \EndIf
        \State Obtain $(v,q)=(f(x'),p(x'))$ by oracle evaluation or exact-cache reuse
        \State Append the edits, oracle outcome, and probability to global history $H$
        \State Create child $n'$ of $n$ with state $x'$ and statistics $N=0,W=0$
        \State $r\gets r(x')$ using $\mathcal R_t$ as $\mathcal S_{\mathrm{valid}}$ in the novelty term
        \State $\operatorname{Backpropagate}(n',r)$ \Comment{Includes oracle-negative children}
        \If{$v=1$}
            \State $\mathcal V_t\gets\mathcal V_t\cup\{x'\}$
        \EndIf
    \EndFor
    \State $\mathcal S\gets\operatorname{ExactUnique}(\mathcal S\cup\mathcal V_t)$
\EndFor
\State \Return $\mathcal S$
\end{algorithmic}
}
\end{algorithm}

\pruningrev{A node with no children can be retried. At $D_{\max}=5$, selection stops at depth four even when children exist, allowing repeated expansion without generating states deeper than five. History $H$ retains records from all branches; only the prompt subset is truncated. Missing and unparsable slots are logged but omitted from $H$. If probabilities are unavailable, $p(x')=f(x')$.}
\FloatBarrier

\clearpage
\section{Additional Experimental Details and Results}
\label{app:new_experiments}

\subsection{ReCo-MCTS Settings and Evaluation Records}
\label{app:experiment_settings}

The saved condition identifiers map to $(K,L)$ as follows:
\texttt{MK1}=$(1,3)$, \texttt{A1}=$(5,1)$,
\texttt{M10}=$(5,3)$, and \texttt{A5}=$(5,5)$.
The name \texttt{M10} is a legacy identifier: it does not mean that the
prompt is limited to ten history records. The four conditions share
the token-bounded history rule in Section~\ref{sec:experimental_setup}.
Archived records are retained even when they no longer fit the prompt.
$L=3$ is the reference condition retained from the existing evaluation,
not a setting newly selected using a distance-and-sparsity threshold.

The fixed BALANCED reward is
\begin{equation}
 r(c)=g(p)\frac{p+0.5P_c+0.5S_c+0.2N_c}{2.2},\qquad
 g(p)=\operatorname{sigmoid}\left(\frac{p-0.5}{0.1}\right),
\end{equation}
where $p$ is the target-class probability, $P_c=1/(1+d(x,c))$,
$S_c=1/(1+s(x,c))$, and
$N_c=1-1/(1+d_{\min}(c))$ for the distance to the nearest previously
found exact-unique valid candidate. $N_c=1$ when this reference set
is empty. All candidates in one call use the reference set from before
that call. \pruningrev{Missing, unparsable, or constraint-violating proposal slots receive zero reward.}
The reward's $S_c$ is a transformed score, whereas the reported S is
the untransformed number of changed features. Its history-dependent
$N_c$ also differs from the final-set Novelty D below.
The reward, history, and UCT are fixed settings in these experiments;
the $K,L$ comparisons do not separately establish their effectiveness.

All ReCo-MCTS runs reach 30 calls. Each model contributes 1,600 runs
(four conditions $\times$ four datasets $\times$ 100 inputs);
the baseline study contributes 3,600 runs (three methods $\times$ three
budgets $\times$ 400 inputs). We reuse the same evaluation inputs
previously examined with Qwen. They are not a newly untouched test set.
GLM and Qwen differ in their deployed model and numerical precision;
their comparison does not isolate model size or quantization.
The supplied final-output audit contains one record for every ReCo-MCTS run,
including empty outputs, and verifies the terminal archive contract.
The baseline audit matches native returns to all 3,600 run summaries.
No candidates were generated and no oracle inference was performed
for the present table construction or GLM statistical analysis.

\subsection{Metric Definitions and Aggregation}
\label{app:evaluation_details}

\paragraph{Distance and changes.}
Let $\mathcal N$ and $\mathcal A$ denote the numerical and categorical
features in the frozen evaluator. Numerical distance is
$d_{\mathcal N}(a,b)=\sqrt{|\mathcal N|^{-1}
\sum_{j\in\mathcal N}((a_j-b_j)/\sigma_j)^2}$, and categorical distance
is $d_{\mathcal A}(a,b)=|\mathcal A|^{-1}
\sum_{j\in\mathcal A}\mathbb{1}[a_j\ne b_j]$.
Their arithmetic mean defines $d$ when both groups are present;
otherwise the nonempty group alone defines it.
$\sigma_j$ is the training median absolute deviation (MAD).
If MAD is zero, the evaluator uses the feature's specified numerical
unit, or $10^{-6}$ when no unit is specified. Thus some numerical
features can contribute very large distances, particularly on Adult.
The change count uses the frozen schema's comparison rule after type
and unit normalization, excluding identifiers and the prediction target.

\paragraph{Representative sets.}
First, repeated identical states are counted once, so repeatedly returning
the same alternative cannot increase the count. For the primary result,
each ordinary numerical value is assigned to a training-defined quantile
bin. States with the same bin indices for all numerical features and
the same categorical values belong to one group. Special sentinel
values have separate keys. For $q$ bins, boundaries are the training
quantiles at $1/q,\ldots,(q-1)/q$, with linear interpolation; duplicate
boundaries are merged. Equality to a boundary goes to the upper bin.
Consequently, features with repeated or constant values can have fewer
than $q$ nonempty bins. Boundaries are fixed using the oracle's training
split, excluding special values, and are shared by all methods.
They are not fitted to generated CFs or chosen to improve a method's result.

The first valid state in each group is its representative, following the
recorded archive or return order. P, S, and the supplementary metrics use
this representative's actual values, not the bin center.
Ten-quantile aggregation limits inflation from small numerical variations
but does not make the count independent of resolution. It is a reproducible
evaluation convention, not a unique definition of distinct recourse.
Exact-state and 5/20-quantile results therefore accompany the primary result.

\paragraph{Averaging and missing values.}
Each successful input contributes one within-set mean to a quality average,
regardless of how many CFs it has. Empty sets have no defined quality;
they contribute zero CFs and still contribute to success and cost summaries.
The four-dataset mean is the mean of the four dataset means, not a pooled
mean of successful inputs. If any dataset has no eligible quality inputs,
the overall quality mean is unavailable. For a paired quality difference,
we restrict each dataset to input IDs that succeeded in both conditions
and average their per-input differences. Thus subtracting two displayed
unpaired quality means need not reproduce the paired estimate.

\paragraph{Supplementary quality.}
For $|C(x)|>1$, Novelty is
\begin{equation}
 D(x)=\frac{1}{|C(x)|}\sum_{c\in C(x)}
          \min_{c'\in C(x),\,c'\ne c}d(c,c').
\end{equation}
We set D to zero for a singleton and leave it undefined for an empty set.
It describes within-set nearest-neighbor separation; it depends on set
size as well as geometry, so a larger D is not by itself evidence of
a better set of recourse options. We do not subsample sets to a common size.

For the distribution-related measures, the reference set consists of
exact-unique rows from the oracle's training split, normalized by the
same frozen schema. NN5 is the mean distance from a CF to its five nearest
reference rows; yNN5 is the fraction of those neighbors classified by
the oracle as the target class. Distance ties use fixed reference IDs,
and an exact match remains a zero-distance neighbor. CF values are first
averaged within an input and then across successful inputs. Reference
labels are used for offline evaluation, not as extra search feedback.
These measures were added after examining the initial Qwen results.
They describe proximity to the reference and local oracle-label agreement,
respectively; neither establishes causal feasibility.

\subsection{Baseline Implementations and Access}
\label{app:baseline_settings}

\paragraph{DiCE.}
We use the unpatched random implementation in DiCE 0.12
\citep{dice_code}. Each restart requests five CFs with
\texttt{sample\_size=1000}, stopping threshold 0.5,
\texttt{posthoc\_sparsity\_param=0.1}, linear post-hoc sparsity,
and a 10,000-step linear-search limit. A restart-specific seed is passed
explicitly. The adapter supplies probabilities and private schema/MAD
metadata, not training rows. The desired class is the common target,
and the common constraint validator applies to all outputs.

\paragraph{Growing Spheres.}
We use the authors' search implementation\footnote{\url{https://github.com/thibaultlaugel/growingspheres}}
at commit \texttt{894ce969f39f}, with 2,000 samples per layer, ring layers,
initial radius 0.1, radius reduction factor 10, and sparse refinement.
The mixed-feature adapter searches editable coordinates, decodes numerical
units and categorical one-hot values, constructs derived features, and
enforces the common constraints. It uses oracle class labels and
training-range scaling. The results apply to this adapted implementation.

\paragraph{C-CHVAE.}
We use CARLA's PyTorch implementation\footnote{\url{https://github.com/carla-recourse/CARLA}}
at commit \texttt{24db00aa8616}, rather than the original authors' TensorFlow
implementation. The mixed-feature adapter handles multiple categories,
numerical units, immutable and sentinel features, and the common oracle.
The search uses 300 samples per step, $\ell_1$ distance, step size 0.1,
up to 1,000 iterations, and clamping.
The VAE uses layer sizes $[512,256,8]$, KL weight 0.3, regularization
$10^{-6}$, learning rate 0.001, batch size 32, and five training epochs.
It accesses the oracle's training rows and class labels, without classifier
gradients. Thus the methods do not have identical data access, and these
results do not establish the performance of every possible C-CHVAE
implementation or training configuration.

\paragraph{Oracle accounting and output contract.}
Every method uses the same constraint and oracle wrapper.
\experimentrev{The budget counts individual candidate evaluations by the
classifier, including candidates in batches and postprocessing.}
A repeated exact input can reuse its stored prediction;
requests and actual evaluations are reported separately.
Native returns are accumulated across restarts until the budget or
termination rule stops the run. If a budget stop interrupts a native
return, internal candidates from that unfinished return are not added
to the returned set. The ReCo-MCTS terminal archive, in contrast, is its
defined final set of accumulated valid states. These distinct output
contracts are material to the returned-set comparison.

\subsection{Statistical Comparisons}
\label{app:returned_statistics}

All GLM statistics here are recomputed from the saved returned-set input
metrics, with analysis seed 20260926. For each comparison, bootstrap
resampling preserves the left/right pairing within a dataset. Each
replicate of the overall estimate averages the four resampled dataset
means. We use 10,000 replicates and percentile intervals.
For zero eligible pairs the estimate is unavailable; for one pair
only the point estimate is reported. The overall interval requires
at least two eligible pairs in each dataset.
Intervals are individual descriptive intervals, not simultaneous intervals.
They quantify input sampling uncertainty conditional on the frozen
oracles and generation seed 42; they do not quantify run-to-run variation.

For the GLM reference condition's CF count, we additionally report
100,000 two-sided paired sign-flip draws against each of three baselines
at each of three budgets. The nine comparisons form one Holm-adjusted
family. A Monte Carlo p-value is $(b+1)/(100{,}000+1)$, where $b$ is
the number of randomized absolute differences at least as large as
observed. The procedure relies on sign exchangeability of paired
differences under the null. With 100 inputs per dataset, the pooled
CF-count difference equals the equal-dataset mean.
All nine comparisons have zero exceedances in these draws, giving
raw $p\approx10^{-5}$ and Holm-adjusted $p\approx9\times10^{-5}$;
these are finite Monte Carlo resolutions, not zero p-values.
The return-contract and model-presentation choices remain post hoc
despite the correction. Earlier discovery-set tests are not reused.

\begin{table}[t]
\centering
\small
\caption{Exploratory GLM ReCo-MCTS ($K=5,L=3$) minus baseline CF counts. Differences and paired 95\% bootstrap intervals average four datasets equally (400 paired inputs). Two-sided sign-flip p-values have add-one correction and Holm adjustment across these nine comparisons.}
\label{tab:glm_count_tests}
\begin{fittoiclr}
\begin{tabular}{llrrrr}
\toprule
Baseline & $B$ & $\Delta$ CF & 95\% CI & $p$ & $p_{
m Holm}$ \\
\midrule
DiCE & 150 & 75.20 & [72.65, 77.72] & 0.00001 & 0.00009 \\
DiCE & 1,000 & 75.20 & [72.66, 77.80] & 0.00001 & 0.00009 \\
DiCE & 10,000 & 54.16 & [51.76, 56.50] & 0.00001 & 0.00009 \\
Growing Spheres & 150 & 75.20 & [72.62, 77.65] & 0.00001 & 0.00009 \\
Growing Spheres & 1,000 & 75.20 & [72.59, 77.75] & 0.00001 & 0.00009 \\
Growing Spheres & 10,000 & 75.17 & [72.66, 77.75] & 0.00001 & 0.00009 \\
\experimentrev{C-CHVAE} & 150 & 75.20 & [72.70, 77.71] & 0.00001 & 0.00009 \\
\experimentrev{C-CHVAE} & 1,000 & 75.20 & [72.65, 77.66] & 0.00001 & 0.00009 \\
\experimentrev{C-CHVAE} & 10,000 & 75.19 & [72.53, 77.75] & 0.00001 & 0.00009 \\
\bottomrule
\end{tabular}
\end{fittoiclr}
\end{table}

\begin{table}[t]
\centering
\small
\caption{GLM CF-count differences between $(K,L)$ settings. Intervals are individual paired 95\% bootstrap intervals, not simultaneous intervals or generation-seed uncertainty.}
\label{tab:glm_setting_effects}
\begin{fittoiclr}
\begin{tabular}{llrr}
\toprule
Left minus right & Dataset & $\Delta$ CF & 95\% CI \\
\midrule
$(5,3)-(1,3)$ & Loan & 68.00 & [62.85, 73.10] \\
 & Adult & 45.16 & [41.99, 48.30] \\
 & Credit & 89.59 & [83.89, 95.04] \\
 & HELOC & 68.27 & [62.74, 73.72] \\
 & Equal mean & 67.75 & [65.28, 70.21] \\
\midrule
$(5,3)-(5,1)$ & Loan & 6.88 & [1.35, 12.60] \\
 & Adult & 19.97 & [16.90, 23.04] \\
 & Credit & 38.07 & [31.12, 44.91] \\
 & HELOC & 25.88 & [21.71, 30.10] \\
 & Equal mean & 22.70 & [20.15, 25.31] \\
\midrule
$(5,5)-(5,1)$ & Loan & 2.39 & [-2.73, 7.56] \\
 & Adult & 19.24 & [16.37, 22.15] \\
 & Credit & 34.29 & [27.32, 41.07] \\
 & HELOC & 20.27 & [15.94, 24.50] \\
 & Equal mean & 19.05 & [16.56, 21.52] \\
\midrule
$(5,5)-(5,3)$ & Loan & -4.49 & [-10.00, 0.95] \\
 & Adult & -0.73 & [-3.30, 1.91] \\
 & Credit & -3.78 & [-8.14, 0.61] \\
 & HELOC & -5.61 & [-9.41, -1.80] \\
 & Equal mean & -3.65 & [-5.71, -1.57] \\
\bottomrule
\end{tabular}
\end{fittoiclr}
\end{table}

\begin{table}[t]
\centering
\small
\caption{GLM ReCo-MCTS minus DiCE ($B=10{,}000$), restricted to the same successful inputs in each pair. Positive $\Delta$P favors ReCo-MCTS; negative $\Delta$S favors ReCo-MCTS. Brackets give individual 95\% bootstrap intervals. The equal mean weights dataset means, not individual successful inputs.}
\label{tab:glm_dice_quality_effects}
\begin{fittoiclr}
\begin{tabular}{llrrr}
\toprule
ReCo-MCTS & Dataset & Paired $n$ & $\Delta$P [95\% CI] & $\Delta$S [95\% CI] \\
\midrule
$L=3$ & Loan & 100 & -0.009 [-0.019, 0.000] & 1.929 [1.874, 1.983] \\
 & Adult & 100 & 0.027 [0.012, 0.045] & 1.971 [1.890, 2.053] \\
 & Credit & 98 & 0.371 [0.330, 0.411] & 2.220 [2.050, 2.377] \\
 & HELOC & 96 & -0.027 [-0.043, -0.012] & 2.643 [2.546, 2.732] \\
 & Equal mean & 394 & 0.090 [0.078, 0.102] & 2.191 [2.138, 2.243] \\
\midrule
$L=1$ & Loan & 100 & -0.040 [-0.050, -0.030] & 0.758 [0.717, 0.797] \\
 & Adult & 100 & 0.097 [0.075, 0.120] & 0.671 [0.628, 0.715] \\
 & Credit & 91 & 0.411 [0.365, 0.455] & 0.836 [0.713, 0.946] \\
 & HELOC & 96 & 0.040 [0.023, 0.056] & 0.827 [0.719, 0.922] \\
 & Equal mean & 387 & 0.127 [0.114, 0.140] & 0.773 [0.729, 0.813] \\
\bottomrule
\end{tabular}
\end{fittoiclr}
\end{table}

\FloatBarrier

\subsection{All Baseline Budgets}
\label{app:baseline_budgets}

\begin{table}[t]
\centering
\small
\caption{All baseline budgets: mean returned CF count / $n_+$ (out of 100). Zero denotes no valid native return before termination, not necessarily no internal valid candidate.}
\label{tab:baseline_budgets}
\begin{fittoiclr}
\begin{tabular}{llrrrr}
\toprule
Method & $B$ & Loan & Adult & Credit & HELOC \\
\midrule
DiCE & 150 & 0.00 / 0 & 0.00 / 0 & 0.00 / 0 & 0.00 / 0 \\
DiCE & 1,000 & 0.00 / 0 & 0.00 / 0 & 0.00 / 0 & 0.00 / 0 \\
DiCE & 10,000 & 24.57 / 100 & 16.65 / 100 & 19.05 / 98 & 23.88 / 100 \\
\midrule
Growing Spheres & 150 & 0.00 / 0 & 0.00 / 0 & 0.00 / 0 & 0.00 / 0 \\
Growing Spheres & 1,000 & 0.00 / 0 & 0.00 / 0 & 0.00 / 0 & 0.00 / 0 \\
Growing Spheres & 10,000 & 0.00 / 0 & 0.02 / 2 & 0.02 / 2 & 0.09 / 8 \\
\midrule
\experimentrev{C-CHVAE} & 150 & 0.00 / 0 & 0.00 / 0 & 0.00 / 0 & 0.00 / 0 \\
\experimentrev{C-CHVAE} & 1,000 & 0.00 / 0 & 0.01 / 1 & 0.00 / 0 & 0.00 / 0 \\
\experimentrev{C-CHVAE} & 10,000 & 0.00 / 0 & 0.04 / 4 & 0.00 / 0 & 0.00 / 0 \\
\bottomrule
\end{tabular}
\end{fittoiclr}
\end{table}

At $B=150$ and $B=1{,}000$, DiCE and Growing Spheres return no
valid CFs. DiCE's native batched search is interrupted before it returns
under these caps. This reflects the interaction of budget enforcement
with the native output contract; it is not evidence that every evaluated
candidate was invalid. C-CHVAE returns one valid CF for one Adult input
at $B=1{,}000$. At $B=10{,}000$, Growing Spheres and C-CHVAE still
have very low returned-set coverage. We retain these results, while
restricting conclusions to the tested implementations, adapters,
constraints, and stopping rules.
\FloatBarrier

\subsection{Qwen Results}
\label{app:qwen_results}

\begin{table}[t]
\centering
\small
\caption{ReCo-MCTS candidate settings with Qwen. All runs use 30 LLM calls. CF is the mean valid CF count after ten-quantile aggregation; $n_+$ is the number of successful inputs out of 100. Bold marks the largest CF count per dataset.}
\label{tab:qwen_settings}
\begin{fittoiclr}
\begin{tabular}{lrrrrrr}
\toprule
Dataset & $K$ & $L$ & CF $\uparrow$ & P $\uparrow$ & S $\downarrow$ & $n_+$ \\
\midrule
Loan & 1 & 3 & 3.15 & 0.688 & 1.99 & 100 \\
 & 5 & 1 & 27.96 & 0.667 & 2.25 & 100 \\
 & 5 & 3 & \textbf{36.53} & 0.644 & 3.09 & 100 \\
 & 5 & 5 & 33.53 & 0.644 & 3.04 & 100 \\
\midrule
Adult & 1 & 3 & 1.25 & 0.209 & 2.94 & 63 \\
 & 5 & 1 & 12.40 & 0.286 & 2.37 & 79 \\
 & 5 & 3 & \textbf{16.21} & 0.169 & 3.29 & 89 \\
 & 5 & 5 & 15.66 & 0.122 & 3.39 & 94 \\
\midrule
Credit & 1 & 3 & 1.84 & 0.833 & 3.06 & 68 \\
 & 5 & 1 & 21.11 & 0.844 & 2.97 & 76 \\
 & 5 & 3 & \textbf{34.44} & 0.802 & 4.14 & 87 \\
 & 5 & 5 & 19.68 & 0.793 & 4.10 & 85 \\
\midrule
HELOC & 1 & 3 & 0.14 & 0.396 & 2.99 & 5 \\
 & 5 & 1 & 2.08 & 0.430 & 2.67 & 17 \\
 & 5 & 3 & \textbf{2.16} & 0.496 & 4.17 & 15 \\
 & 5 & 5 & 0.59 & 0.482 & 4.29 & 13 \\
\bottomrule
\end{tabular}
\end{fittoiclr}
\end{table}

The same $K,L$ settings are evaluated on Qwen3.6-35B-A3B.
At $L=3$, $K=5$ improves the equal-dataset CF count from 1.60 to 22.34.
For $K=5$, counts are 15.89, 22.34, and 17.37 at $L=1,3,5$,
respectively. The quantity trend agrees with GLM, but absolute coverage
is substantially lower: the Qwen reference condition succeeds on 291/400
inputs, including only 15/100 on HELOC, versus 395/400 and 96/100 for GLM.
Qwen's reference CF count is below DiCE on Adult and HELOC,
whereas it is above DiCE on Loan and Credit.
Its overall advantage over DiCE at ten-quantile resolution is only
1.30 CFs. The supplied returned-set analysis reports a paired interval
$[-0.32,2.91]$ and Holm-adjusted $p=0.2392$ for this Qwen comparison.
Those values belong to the separate nine-comparison Qwen family;
they are not GLM statistics or part of a joint cross-model correction.
The baseline values to use for these comparisons are in
Table~\ref{tab:glm_baselines}.

With Qwen, $L=1$ uses fewer changed features than $L=3$ on every dataset.
It also improves proximity on Loan, Adult, and Credit, while HELOC has
higher P at $L=3$. The exception differs from GLM, reinforcing that the
proximity trade-off is model- and dataset-dependent.
\FloatBarrier

\subsection{Supplementary Quality}
\label{app:supplementary_quality}

\begin{table}[t]
\centering
\small
\caption{GLM and baselines ($B=10{,}000$). Supplementary quality on ten-quantile representatives. D is within-set nearest-neighbor distance; NN5 and yNN5 use the frozen training reference. Interpret distances within each dataset. $n_+$ is out of 100.}
\label{tab:glm_quality}
\begin{fittoiclr}
\begin{tabular}{llrrrr}
\toprule
Dataset & Setting / method & Novelty D & NN5 $\downarrow$ & yNN5 $\uparrow$ & $n_+$ \\
\midrule
Loan & $K=1,L=3$ & 0.093 & 0.316 & 0.975 & 100 \\
 & $K=5,L=1$ & 0.090 & 0.372 & 0.941 & 100 \\
 & $K=5,L=3$ & 0.070 & 0.332 & 0.912 & 100 \\
 & $K=5,L=5$ & 0.071 & 0.333 & 0.912 & 100 \\
 & DiCE & 0.181 & 0.367 & 0.795 & 100 \\
 & Growing Spheres & -- & -- & -- & 0 \\
 & \experimentrev{C-CHVAE} & -- & -- & -- & 0 \\
\midrule
Adult & $K=1,L=3$ & 182.254 & 130.346 & 0.920 & 100 \\
 & $K=5,L=1$ & 95.920 & 196.643 & 0.920 & 100 \\
 & $K=5,L=3$ & 40.165 & 151.948 & 0.903 & 100 \\
 & $K=5,L=5$ & 57.845 & 130.313 & 0.908 & 100 \\
 & DiCE & 842.182 & 1920.335 & 0.737 & 100 \\
 & Growing Spheres & 0.000 & 4.990 & 0.400 & 2 \\
 & \experimentrev{C-CHVAE} & 0.000 & 102.759 & 0.850 & 4 \\
\midrule
Credit & $K=1,L=3$ & 0.417 & 0.718 & 0.741 & 96 \\
 & $K=5,L=1$ & 0.269 & 1.358 & 0.667 & 91 \\
 & $K=5,L=3$ & 0.145 & 0.714 & 0.742 & 99 \\
 & $K=5,L=5$ & 0.139 & 0.594 & 0.736 & 99 \\
 & DiCE & 9.448 & 8.449 & 0.871 & 98 \\
 & Growing Spheres & 0.000 & 0.235 & 0.100 & 2 \\
 & \experimentrev{C-CHVAE} & -- & -- & -- & 0 \\
\midrule
HELOC & $K=1,L=3$ & 0.295 & 1.582 & 0.383 & 91 \\
 & $K=5,L=1$ & 0.325 & 1.389 & 0.419 & 96 \\
 & $K=5,L=3$ & 0.277 & 1.473 & 0.484 & 96 \\
 & $K=5,L=5$ & 0.282 & 1.518 & 0.483 & 100 \\
 & DiCE & 1.504 & 2.146 & 0.322 & 100 \\
 & Growing Spheres & 0.013 & 1.221 & 0.537 & 8 \\
 & \experimentrev{C-CHVAE} & -- & -- & -- & 0 \\
\bottomrule
\end{tabular}
\end{fittoiclr}
\end{table}

\begin{table}[t]
\centering
\small
\caption{Qwen. Supplementary quality on ten-quantile representatives. D is within-set nearest-neighbor distance; NN5 and yNN5 use the frozen training reference. Interpret distances within each dataset. $n_+$ is out of 100.}
\label{tab:qwen_quality}
\begin{fittoiclr}
\begin{tabular}{llrrrr}
\toprule
Dataset & Setting / method & Novelty D & NN5 $\downarrow$ & yNN5 $\uparrow$ & $n_+$ \\
\midrule
Loan & $K=1,L=3$ & 0.058 & 0.315 & 0.976 & 100 \\
 & $K=5,L=1$ & 0.068 & 0.332 & 0.939 & 100 \\
 & $K=5,L=3$ & 0.052 & 0.327 & 0.969 & 100 \\
 & $K=5,L=5$ & 0.053 & 0.326 & 0.961 & 100 \\
\midrule
Adult & $K=1,L=3$ & 563.590 & 80.596 & 0.900 & 63 \\
 & $K=5,L=1$ & 256.879 & 143.983 & 0.866 & 79 \\
 & $K=5,L=3$ & 174.788 & 161.583 & 0.869 & 89 \\
 & $K=5,L=5$ & 154.726 & 164.988 & 0.882 & 94 \\
\midrule
Credit & $K=1,L=3$ & 0.090 & 0.397 & 0.659 & 68 \\
 & $K=5,L=1$ & 0.119 & 0.449 & 0.617 & 76 \\
 & $K=5,L=3$ & 0.079 & 0.430 & 0.598 & 87 \\
 & $K=5,L=5$ & 0.087 & 0.414 & 0.620 & 85 \\
\midrule
HELOC & $K=1,L=3$ & 0.744 & 1.737 & 0.360 & 5 \\
 & $K=5,L=1$ & 0.412 & 1.527 & 0.555 & 17 \\
 & $K=5,L=3$ & 0.152 & 1.478 & 0.439 & 15 \\
 & $K=5,L=5$ & 0.378 & 1.469 & 0.463 & 13 \\
\bottomrule
\end{tabular}
\end{fittoiclr}
\end{table}

Tables~\ref{tab:glm_quality} and~\ref{tab:qwen_quality} retain ordinary
Novelty D and the two reference-neighbor measures without adding a
fixed-size diversity metric. At the reference GLM setting, D is lower
than DiCE's on all datasets, while NN5 is also lower on all datasets.
yNN5 is higher for GLM on Loan, Adult, and HELOC, but lower on Credit.
Thus the larger GLM sets have smaller nearest-neighbor separation,
and their closer training-reference neighbors do not always yield
stronger target-label agreement. These are complementary observations,
not a uniform quality improvement. Adult's large D and NN5 values
reflect the frozen distance scale, including its zero-MAD fallback;
we report them by dataset instead of ranking methods by a pooled
distance score. Very small successful subsets for the other baselines
also limit interpretation of their quality means.
\FloatBarrier

\subsection{Sensitivity to CF-count Resolution}
\label{app:count_sensitivity}

\begin{table}[t]
\centering
\small
\caption{CF-count sensitivity to aggregation ($K=5$ for ReCo-MCTS; $B=10{,}000$ for DiCE). Exact merges identical states only; 5/10/20 use training quantile boundaries. All other conditions are in the analysis CSV.}
\label{tab:count_sensitivity}
\begin{fittoiclr}
\begin{tabular}{llrrrr}
\toprule
Dataset & Setting / method & Exact & 5 & 10 & 20 \\
\midrule
Loan & GLM ($L=3$) & 120.30 & 41.87 & 76.61 & 100.20 \\
 & GLM ($L=1$) & 116.48 & 41.54 & 69.73 & 88.27 \\
 & Qwen ($L=3$) & 99.41 & 18.52 & 36.53 & 61.73 \\
 & Qwen ($L=1$) & 60.07 & 15.76 & 27.96 & 37.83 \\
 & DiCE & 33.26 & 19.43 & 24.57 & 28.30 \\
\midrule
Adult & GLM ($L=3$) & 117.53 & 42.67 & 51.36 & 68.19 \\
 & GLM ($L=1$) & 86.30 & 29.01 & 31.39 & 41.19 \\
 & Qwen ($L=3$) & 27.92 & 15.32 & 16.21 & 19.04 \\
 & Qwen ($L=1$) & 20.78 & 12.25 & 12.40 & 13.68 \\
 & DiCE & 49.71 & 15.97 & 16.65 & 19.58 \\
\midrule
Credit & GLM ($L=3$) & 109.00 & 89.40 & 99.92 & 105.04 \\
 & GLM ($L=1$) & 65.72 & 57.86 & 61.85 & 63.75 \\
 & Qwen ($L=3$) & 35.50 & 33.99 & 34.44 & 34.76 \\
 & Qwen ($L=1$) & 23.71 & 20.41 & 21.11 & 21.32 \\
 & DiCE & 24.81 & 17.97 & 19.05 & 19.48 \\
\midrule
HELOC & GLM ($L=3$) & 94.75 & 54.09 & 72.91 & 79.22 \\
 & GLM ($L=1$) & 57.74 & 38.10 & 47.03 & 49.58 \\
 & Qwen ($L=3$) & 3.94 & 1.23 & 2.16 & 2.92 \\
 & Qwen ($L=1$) & 3.56 & 1.47 & 2.08 & 2.57 \\
 & DiCE & 34.75 & 18.73 & 23.88 & 27.70 \\
\bottomrule
\end{tabular}
\end{fittoiclr}
\end{table}

For GLM at $K=5,L=3$, the mean CF counts under exact, 5-, 10-, and
20-quantile aggregation are 110.40, 57.01, 75.20, and 88.16.
DiCE at $B=10{,}000$ gives 35.63, 18.03, 21.04, and 23.77,
respectively. GLM's count advantage holds on every dataset under all
four conventions. The magnitude remains resolution-dependent.
Qwen's weaker advantage is less robust: at five-quantile resolution,
its reference mean is 17.27 versus 18.03 for DiCE, and its Loan count
is 18.52 versus 19.43. We therefore do not generalize the GLM
baseline comparison to every model or aggregation rule.
\FloatBarrier

\subsection{Recorded Search Costs}
\label{app:search_costs}

\begin{table}[t]
\centering
\small
\caption{Mean recorded search cost per input, averaged equally across four datasets. Requests include cache hits; \experimentrev{evaluations count candidates actually evaluated by the classifier}. Tokens include input and output. Search time excludes model startup, VAE training and offline evaluation and is not hardware-normalized.}
\label{tab:search_costs}
\begin{fittoiclr}
\begin{tabular}{llrrrrr}
\toprule
Model / method & Setting & Calls & Requests & Evaluations & Tokens (k) & Time (s) \\
\midrule
GLM & $K=1,L=3$ & 30 & 28.4 & 17.5 & 70.2 & 58.51 \\
GLM & $K=5,L=1$ & 30 & 146.4 & 121.3 & 82.0 & 159.03 \\
GLM & $K=5,L=3$ & 30 & 139.5 & 126.6 & 87.9 & 232.15 \\
GLM & $K=5,L=5$ & 30 & 135.7 & 122.5 & 91.0 & 268.44 \\
\midrule
Qwen & $K=1,L=3$ & 30 & 21.9 & 5.1 & 72.0 & 15.38 \\
Qwen & $K=5,L=1$ & 30 & 118.6 & 71.1 & 79.0 & 44.06 \\
Qwen & $K=5,L=3$ & 30 & 80.0 & 65.1 & 84.2 & 72.72 \\
Qwen & $K=5,L=5$ & 30 & 68.2 & 51.6 & 87.0 & 88.99 \\
\midrule
DiCE & $B=150$ & 0 & 1001.0 & 150.0 & 0 & 0.06 \\
DiCE & $B=1,000$ & 0 & 2001.0 & 1000.0 & 0 & 0.11 \\
DiCE & $B=10,000$ & 0 & 17044.7 & 10000.0 & 0 & 1.03 \\
Growing Spheres & $B=150$ & 0 & 2001.0 & 150.0 & 0 & 0.11 \\
Growing Spheres & $B=1,000$ & 0 & 2001.0 & 1000.0 & 0 & 0.11 \\
Growing Spheres & $B=10,000$ & 0 & 142573.5 & 10000.0 & 0 & 8.49 \\
\experimentrev{C-CHVAE} & $B=150$ & 0 & 14675.9 & 144.2 & 0 & 1.33 \\
\experimentrev{C-CHVAE} & $B=1,000$ & 0 & 25949.3 & 926.5 & 0 & 2.31 \\
\experimentrev{C-CHVAE} & $B=10,000$ & 0 & 290333.2 & 9118.7 & 0 & 27.47 \\
\bottomrule
\end{tabular}
\end{fittoiclr}
\end{table}

The cost table separates LLM calls, total tokens, oracle requests,
actual oracle evaluations, and elapsed search time. LLM calls and
\experimentrev{oracle evaluations} are not interchangeable units. Reported
times come from each method's recorded execution environment, not
a controlled hardware-matched timing study. In particular, model
precision and inference configurations differ across GLM and Qwen.
The search times exclude model startup, generative-model training,
and offline metric computation; they should not be interpreted as
total end-to-end deployment costs. The present saved-table analysis
adds no generation or oracle-evaluation cost.
\FloatBarrier

\fi

\end{document}